\documentclass[runningheads]{llncs}

\usepackage{eccv}

\usepackage{eccvabbrv}

\usepackage{graphicx}
\usepackage{booktabs}
\usepackage{array} 
\usepackage[accsupp]{axessibility}  
\usepackage{wrapfig}

\usepackage{hyperref}

\usepackage{orcidlink}

\begin{document}

\title{GroundingAnomaly: Spatially-Grounded Diffusion for Few-Shot Anomaly Synthesis} 

\titlerunning{GroundingAnomaly}
\author{
Yishen Liu\inst{1} \and
Hongcang Chen\inst{1, 2} \and
Pengcheng Zhao\inst{3} \and
Yunfan Bao\inst{1} \and
Yuxi Tian\inst{1} \and
Jieming Zhang \inst{4} \and
Hao Chen \inst{4} \and
Zhi Zheng\inst{4} \and
Yongchun Liu \inst{4} \and
Ying Li\inst{1}\thanks{Corresponding author.} \and
Dongpu Cao \inst{5}\protect\footnotemark[1] }

\authorrunning{Y.~Liu et al.}

\institute{
Beijing Institute of Technology, Beijing, China
\\\email{ying.li@bit.edu.cn}
\and 
Hong Kong Polytechnic University, Hong Kong, China
\and
Beijing Jiaotong University, Beijing, China
\and
Li Auto, China
\and
Tsinghua University, Beijing, China 
\\\email{dongpu.cao@tsinghua.edu.cn}
}

\maketitle

\begin{abstract}
    The performance of visual anomaly inspection in industrial quality control is often constrained by the scarcity of real anomalous samples. Consequently, anomaly synthesis techniques have been developed to enlarge training sets and enhance downstream inspection. However, existing methods either suffer from poor integration caused by inpainting or fail to provide accurate masks. To address these limitations, we propose \textbf{GroundingAnomaly}, a novel few-shot anomaly image generation framework. Our framework introduces a \textbf{Spatial Conditioning Module} that leverages per-pixel semantic maps to enable precise spatial control over the synthesized anomalies. Furthermore, a \textbf{Gated Self-Attention Module} is designed to inject conditioning tokens into a frozen U-Net via gated attention layers. This carefully preserves pretrained priors while ensuring stable few-shot adaptation. Extensive evaluations on the MVTec AD and VisA datasets demonstrate that GroundingAnomaly generates high-quality anomalies and achieves state-of-the-art performance across multiple downstream tasks, including anomaly detection, segmentation, and instance-level detection. Code will be made publicly available from \href{https://github.com/Eeelyyy/GroundingAnomaly}{this URL}.
\end{abstract}

\section{Introduction}
\label{sec:intro}

Recently, visual anomaly inspection has demonstrated significant potential in industrial quality control in manufacturing \cite{cao2024survey}. 
Nevertheless, anomalous samples are scarce in real-world industrial production, which constrains the performance of anomaly inspection. To mitigate this, many methods adopt unsupervised learning on abundant normal samples, detecting anomalies as deviations from the learned distribution \cite{deng2022anomaly, Roth_2022_CVPR, zhang2023exploring, jeong2023winclip, he2024mambaad, li2024promptad, ma2025remp, guo2025one}. However, such approaches exhibit limited localization accuracy and do not provide class-aware anomaly information. 

Such limitation motivates the development of anomaly generation techniques to augment existing datasets. Early model-free methods \cite{Zavrtanik_2021_ICCV, li2021cutpaste} synthesize pseudo-anomalies through data augmentation but suffer from low fidelity and defect appearance remains limited without specific classification. More recently, generative models \cite{goodfellow2014generative, ho2020denoising, song2020denoising, rombach2022high} have been adopted for anomaly synthesis. By leveraging large-scale pretrained priors, these models significantly enhance both visual realism and anomaly diversity. Existing model-based approaches fall into two categories. \textbf{Anomaly Generation (AG)} methods that synthesize isolated defect patches and edit them onto real backgrounds \cite{hu2023anomalydiffusion, gui2024few}, while they preserve background coherence, inpainting often yields poor integration and misaligned masks. \textbf{Anomaly Image Generation (AIG)} methods jointly synthesize objects and defects to improve global realism \cite{Duan2023DFMGAN, jin2025dual, dai2025SeaS}, but they still struggle to produce precise masks.

\begin{figure}[htbp]
  \centering
  \includegraphics[width=0.7\columnwidth,trim=7.5cm 0.5cm 7.5cm 0.5cm,clip]{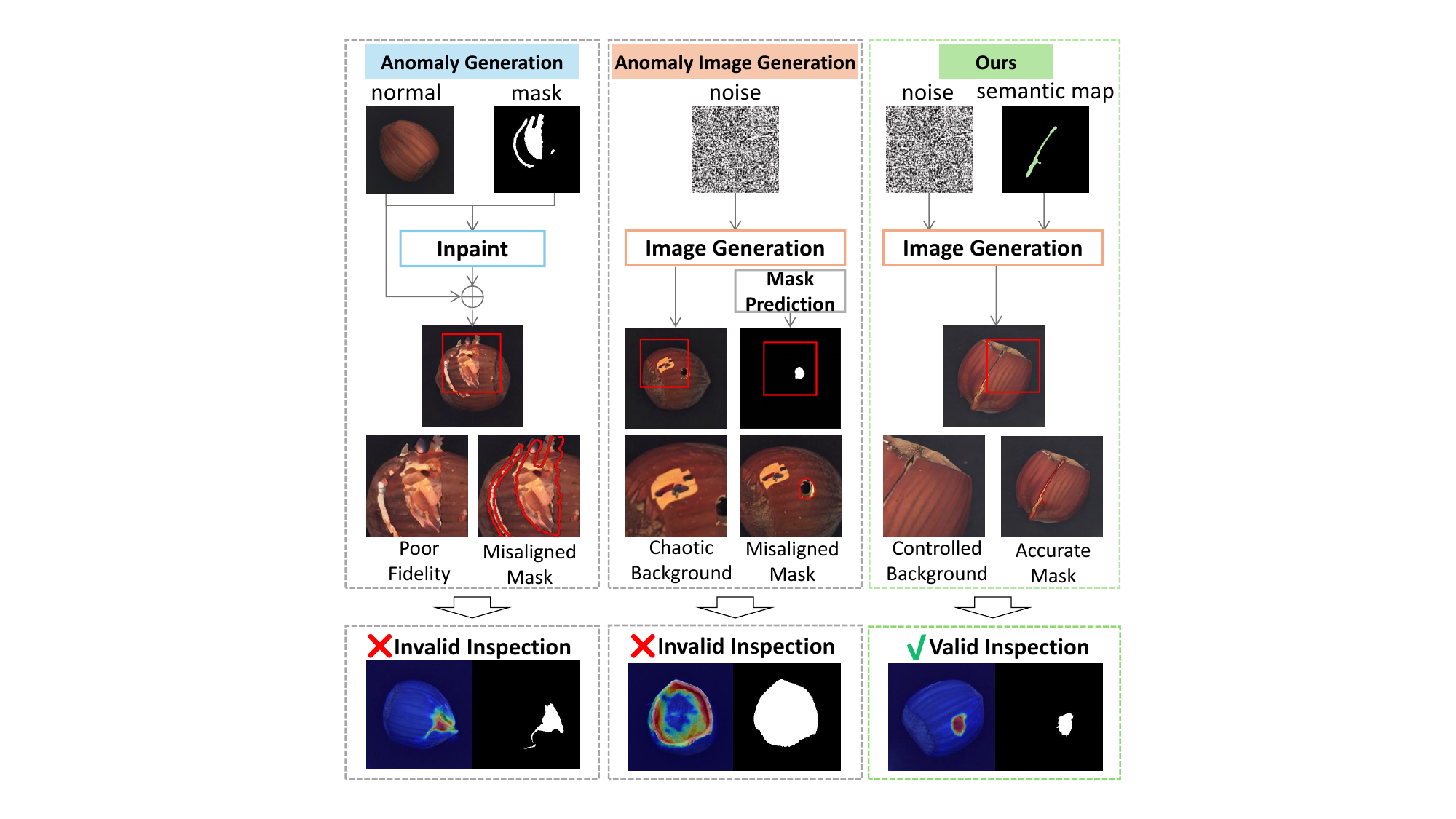}
  \caption{\textbf{Anomaly Generation} methods inpaint anomalies onto normal images; \textbf{Anomaly Image Generation} methods jointly generate anomalies with products and predict masks after generation; our \textbf{GroundingAnomaly} grounds anomalies with semantic maps and generates the whole anomalous images.}
\end{figure}

These limitations motivate us to develop a diffusion framework that provides precise spatial \textbf{grounding} for \textbf{anomaly} image generation while preserving high visual fidelity. In this work, we propose \textbf{GroundingAnomaly}, an AIG framework that learns anomaly appearance and position representation in a few samples and generates realistic anomalous images with precise spatial control. GroundingAnomaly first learns a set of product and anomaly tokens and encodes a pixel-wise semantic map; these are merged in the Spatial Conditioning Module (SCM) to produce conditioning tokens. The conditioning tokens are injected into the diffusion U-Net and fused with the visual tokens in the transformer blocks via a Gated Self-Attention Module (GSM). The modules are trained on mixed batches of normal and anomalous images to leverage cross-domain appearance priors, enabling the model to generate diverse, high-fidelity defects with precise spatial grounding while maintaining global product consistency, which effectively enhances downstream anomaly inspection tasks.

Extensive experiments demonstrate that the proposed GroundingAnomaly generates high-quality anomalies with precise masks and improves downstream anomaly inspection performance reaching a pixel-level 99.3\% AUROC and 85.9\% AP score in anomaly segmentation on MVTec AD \cite{8954181} dataset and 98.2\% AUROC and 67.2\% AP score on VisA \cite{zou2022spot} dataset. 

Our main contributions are summarized as follows:
\begin{itemize}
  \item We propose \textbf{GroundingAnomaly}, a few-shot AIG framework that achieves precise spatial and semantic control over defect synthesis while preserving coherent backgrounds. 
  \item We introduce the \textbf{Spatial Conditioning Module}, which fuses disentangled product and anomaly tokens with a pixel-wise semantic map, and the \textbf{Gated Self-Attention Module}, which injects the spatial conditioning into a frozen U-Net. 
  \item GroundingAnomaly is evaluated on MVTec AD \cite{8954181} and VisA \cite{zou2022spot} using multiple downstream models for anomaly detection, segmentation and instance-level anomaly detection. Experimental results demonstrate state-of-the-art generation quality and substantial improvements in downstream anomaly inspection.
\end{itemize}

\section{Related Works}
\label{sec:rel}

\subsection{Anomaly Inspection}
\noindent \textbf{Anomaly Inspection} is critical for maintaining product quality in modern manufacturing \cite{heckler2025mvtec}. A common approach is to train \textbf{supervised} object-detection or segmentation models on annotated anomalous images to perform instance-level detection or anomaly segmentation \cite{ren2015faster, carion2020end, glenn_jocher_2020_4154370, xiao2018unified, yu2021bisenet, xie2021segformer}. However, real anomalous samples are rare and highly scarce in practice, so supervised approaches often suffer from poor generalization and are constrained by the substantial cost of collecting pixel-accurate annotations. 

\textbf{Unsupervised} methods have been proposed to alleviate this issue. These methods are trained solely on normal examples and detect anomalies by measuring deviations from the learned data distribution \cite{deng2022anomaly, Roth_2022_CVPR, he2024mambaad, zhang2023exploring, guo2025one}. Recent few-shot variants further exploit large-scale pretrained models to perform inspection with only a few exemplars \cite{NEURIPS2022_1d774c11, NEURIPS2023_1abc87c6, Jeong_2023_CVPR, zhou2023anomalyclip, jeong2023winclip, li2024promptad, ma2025remp}. Nevertheless, these approaches only segment anomalies from images without class information, and their performances are still constrained by the representation learned from normal images.

\subsection{Anomaly Synthesis}
\label{sub:as}
\noindent Many anomaly synthesis methods have been developed to generate realistic anomalous data for training inspection models. 
Early model-free methods \cite{Zavrtanik_2021_ICCV, li2021cutpaste} synthesize pseudo-anomalies through data augmentation but suffer from low fidelity and inconsistent defect patterns. 
More recently, generative models \cite{goodfellow2014generative, ho2020denoising, song2020denoising, rombach2022high} have been adopted for anomaly synthesis. By leveraging large-scale pretrained priors, they substantially improve visual realism and diversity.

Existing model-based approaches can be broadly grouped into two categories:
\textbf{Anomaly Generation} methods synthesize isolated defect patches that are composited onto real normal backgrounds. For example, AnomalyDiffusion \cite{hu2023anomalydiffusion} uses Textual Inversion \cite{gal2022image} to learn anomaly appearance and spatial priors and then synthesizes defects within masked regions of normal images. Although they preserve background coherence, such inpainting-based approaches often fail to integrate anomalies seamlessly with the surrounding context and produce misaligned anomaly masks. 

\textbf{Anomaly Image Generation} methods synthesize anomalies together with their host products to ensure coherence and realism. DFMGAN \cite{Duan2023DFMGAN} pretrains a StyleGAN-based generator \cite{Karras2020ada} on normal data and fine-tunes it on a few anomalous exemplars. 
DualAnoDiff \cite{jin2025dual} uses a dual-branch U-Net to separately model normal and anomalous content, significantly improving the fidelity. 
SeaS \cite{dai2025SeaS} finetunes a shared U-Net with unbalanced abnormal 
embeddings to preserve global consistency while maintaining anomaly diversity. 
Despite improving image coherence, existing AIG methods struggle to generate precise masks because they derive masks from upsampled low-resolution attention maps or post-hoc segmentation, which lack pixel-level precision. We propose an AIG framework that generates coherent anomalous images with precise, pixel-aligned spatial grounding.

\subsection{Diffusion Models}
Diffusion models are probabilistic generative models that learn a data distribution \(p_{data}(x)\) by reversing a fixed noising Markov chain of length \(T\). For image synthesis \cite{ho2020denoising, song2020denoising}, U-Net backbones are implemented to predict the noise injected at each timestep, enabling reconstruction of the denoised image.
More recently, Stable Diffusion \cite{rombach2022high} conditions generation via cross-attention to text or other modalities, and ControlNet \cite{zhang2023adding} augments pretrained U-Net with a lightweight, zero-initialized copy to enable precise spatial conditioning. Moreover, GLIGEN \cite{li2023gligen} introduces gated attention layers that inject layout tokens into a pretrained diffusion backbone, enabling controllable layout-to-image generation. Our GroundingAnomaly adopts gated self-attention layers and strengthens spatial grounding and few-shot adaptation via the following designs.

\section{Method}
\label{sec:method}

\begin{figure*}[htb]
  \centering
  \includegraphics[width=1.0\columnwidth,trim=1cm 4.0cm 1cm 3.5cm,clip]{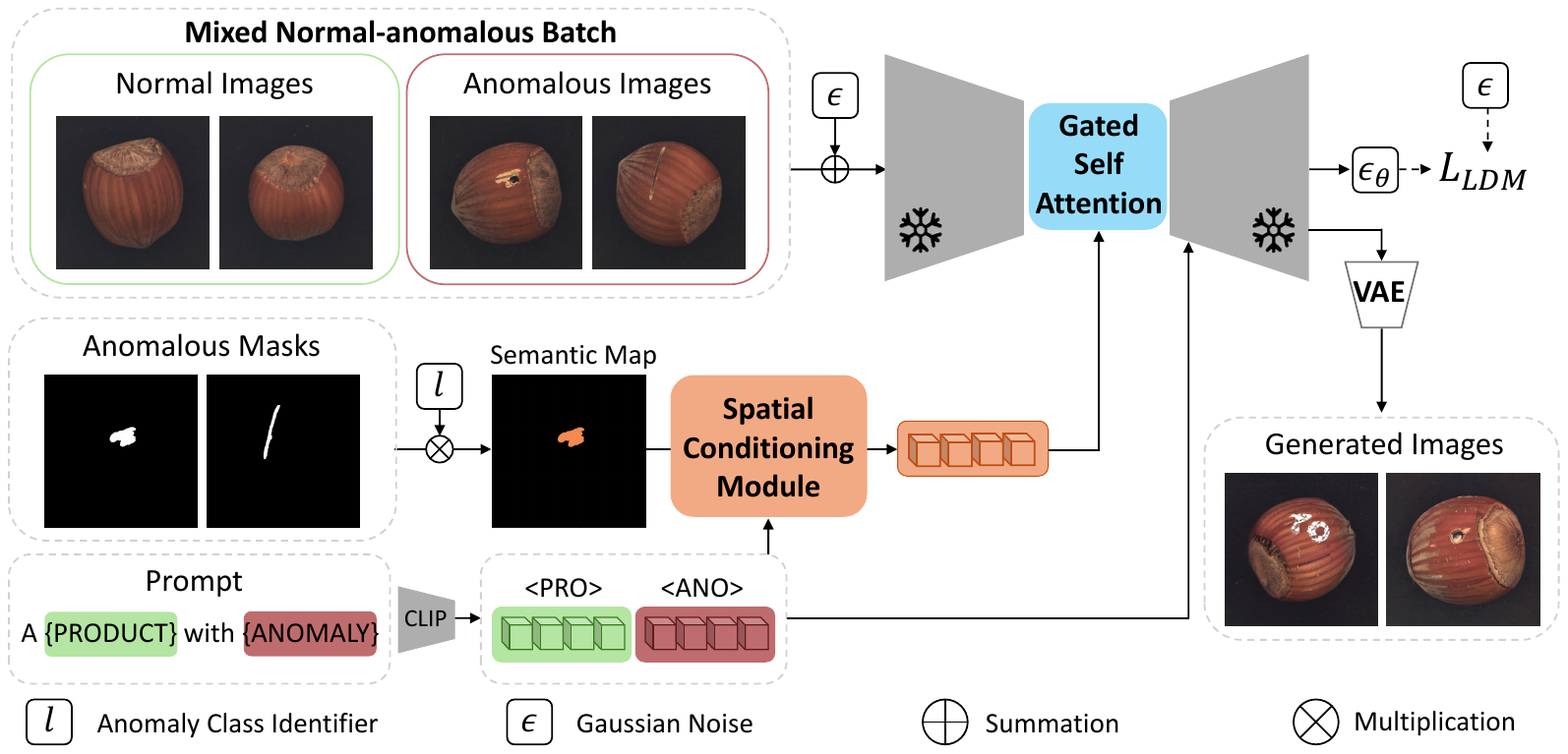}
  \caption{Proposed framework of \textbf{GroundingAnomaly}: (i) \textbf{Spatial Conditioning Module} that encodes a pixel-wise semantic map fuses with disentangled product and anomaly tokens; (ii) \textbf{Gated Self-Attention Module}, which injects the spatial conditioning into a frozen U-Net. (iii) The framework is trained on mixed batches of normal and anomalous images to leverage cross-domain appearance priors, and it generates diverse, high-fidelity anomaly images. }
  \label{fig:framework}
\end{figure*}

\noindent Synthesizing diverse, high-fidelity anomalies with spatially accurate control from limited examples poses two key challenges: (i) learning disentangled appearance and position representations for each anomaly type, and (ii) enabling stable few-shot adaptation without degrading pretrained priors. Our method is based on a pretrained Stable Diffusion v1.4 backbone~\cite{rombach2022high} and addresses both  challenges by introducing two core modules: 
(i) \textbf{Spatial Conditioning Module} (\cref{subsec:spatial}) which encodes a pixel-wise semantic map and fuses it with learned product and anomaly tokens to provide explicit, per-pixel conditioning for appearance and localization control and (ii) \textbf{Gated Self-Attention Module} (\cref{subsec:fuser}) which injects the fused conditioning tokens into the U-Net transformer blocks via gated self-attention, enabling expressive few-shot adaptation while preserving pretrained knowledge. 
Modules are trained on both normal and anomalous images to exploit cross-domain appearance representations between products and defects, thereby improving generation fidelity and diversity; training is performed under the conditional diffusion objective (\cref{eq:ldm}).



\subsection{Preliminaries}
\label{subsec:pre}
\noindent \textbf{Conditioning Diffusion Models.} Latent Diffusion Model \cite{rombach2022high} (LDM) learns data representations in the latent space of a variational auto-encoder (VAE) \cite{kingma2013auto}. 
Given the latent representation \(z_0=\mathcal{E}(x_0)\), 
the forward process injects Gaussian noise \(\epsilon\sim\mathcal{N}(0,I)\) over \(T\) timesteps 
to produce the sequence $\boldsymbol{z}=\{z_1,\ldots,z_T\}$. The reverse denoising process is parameterized by a neural 
network \(\epsilon_\theta(z_t,t)\) trained to predict the injected noise at each timestep, where \(z_t\) denotes the noised latent at timestep \(t\), thereby enabling recovery 
of \(z_0\) from \(z_T\).

Various conditioning strategies such as Stable Diffusion \cite{rombach2022high} have been proposed to guide the denoising process. Let \(\mathbf{v}=\{v_1,\dots,v_M\}\) denote the visual feature tokens of an image in the transformer blocks, the LDM applies residual self-attention on the visual tokens followed by cross-attention with conditioning tokens $\mathbf{h}$, the two residual updates are:
\begin{align}
\mathbf{v} &\leftarrow \mathbf{v} + \mathrm{SelfAttn}(\mathbf{v}), \label{eq:sa}\\
\mathbf{v} &\leftarrow \mathbf{v} + \mathrm{CrossAttn}(\mathbf{v}, \mathbf{h}). \label{eq:ca}
\end{align}

Accordingly, the learning objective of a conditional LDM can be written as
\begin{equation}
\mathcal{L}_{\mathrm{LDM}} \;=\; \mathbb{E}_{x\sim p_{\mathrm{data}},\,\epsilon\sim\mathcal{N}(0,I),\,t}
\left\| \epsilon - \epsilon_\theta(z_t,t,\mathbf{h}) \right\|_2^2,
\label{eq:ldm}
\end{equation}
where \(p_{\mathrm{data}}\) denotes the data distribution of \(x\). 
At inference, sampling begins from \(z_T\sim\mathcal{N}(0,I)\) and the reverse denoising chain produces \(z_0\), 
which is decoded by the VAE to produce the final image.

\subsection{Spatial Conditioning Module}
\label{subsec:spatial}
\noindent\textbf{Disentangled Token Learning.}
To enable few-shot adaptation, we learn compact token embeddings that disentangle anomaly appearance from product identity. Motivated by the observation that anomalies manifest on products rather than as independent entities, we design the prompt template:
\[
\texttt{"A photo of a \{PRODUCT\} with \{ANOMALY\}"},
\]
where \(\{\texttt{PRODUCT}\}\) and \(\{\texttt{ANOMALY}\}\) are placeholders for learnable token sets \(\{\langle\text{pro}_n\rangle\}_{n=1}^N\) and \(\{\langle\text{ano}_k\rangle\}_{k=1}^K\), representing the host product and anomaly class, respectively.
Concretely, we maintain a compact set of  \(N\) product tokens \(\{\langle\text{pro}_n\rangle\}_{n=1}^N\) for each product and \(K\) anomaly tokens \(\{\langle\text{ano}_k\rangle\}_{k=1}^K\) for each anomaly class. 

Unlike Textual Inversion~\cite{gal2022image}, which optimizes a single global token per concept from holistic images, our method learns spatially grounded tokens: each embedding is explicitly aligned with semantic maps, enabling precise control over both anomaly location and local appearance.

\noindent \textbf{Semantic Map Representation.}
Binary masks, while indicating anomaly presence, fail to encode anomaly types. To achieve precise spatial and appearance grounding, we adopt a dense semantic map that jointly encodes anomaly location and class identity. Let \(C\) denote the number of anomaly classes for a given product. We assign each anomaly type a unique identifier \(l\in\{1,\ldots,C\}\) and represent per-pixel semantics by an integer-valued map: 
\begin{equation}
S(x,y)=
\begin{cases}
0, & \text{if pixel }(x,y)\text{ is background};\\[4pt]
l, & \text{if pixel }(x,y)\text{ belongs to anomaly class }l.
\end{cases}
\end{equation}
Where \((x,y)\in\{1,\ldots,H_S\}\times\{1,\ldots,W_S\}\), \(H_S\) and \(W_S\) represent the height and width of the semantic map, respectively.

\noindent \textbf{Spatial-Textual Feature Fusion.}
The semantic map \(S\) is encoded by a pretrained ConvNeXt-Tiny encoder~\cite{liu2022convnet} into a spatial feature map \(F_\mathrm{S}\in\mathbb{R}^{H_f\times W_f\times D_f}\), where \(H_f=W_f=8\) and \(D_f=768\) aligns with the CLIP text embedding space.
To align textual semantics with spatial layout, we construct a textual feature map \(F_\mathrm{T}\) that mirrors the spatial grid of \(F_\mathrm{S}\). Specifically, the learned tokens \(\{\langle\text{pro}_n\rangle\}_{n=1}^N\) and $\{\langle\text{ano}_k\rangle\}_{k=1}^K$ are concatenated into a vector of dimension \(N \!\times\! D_\text{tok}\) (or \(K \!\times\! D_\text{tok}\)) and projected to dimension $D_f$ via a multi-layer perceptron (MLP), then spatially broadcasted according to the textual semantic map $S_f$.
The textual feature map $F_\mathrm{T}\in\mathbb{R}^{H_f\times W_f\times D_f}$ is constructed by assigning each spatial location $(i,j)$ the embedding corresponding to its semantic label in the textual semantic map $S_f$: Let $\tilde{e}_{\mathrm{pro}}$ and $\{\tilde{e}_{\mathrm{ano},l}\}_{l=1}^C$ denote the product and anomaly embeddings after MLP projection to dimension $D_f$. The construction is as follows:
\begin{equation}
F_\mathrm{T}(i,j)=
\begin{cases}
\tilde{e}_{\mathrm{pro}}, & \text{if } S_{f}(i,j)=0; \\
\tilde{e}_{\mathrm{ano}, l}, & \text{if } S_{f}(i,j)=l, \text{ for } l \in \{1, \dots, C\}.
\end{cases}
\end{equation}
The textual map $F_T$ is concatenated with the spatial map $F_S$ along the channel dimension ($D_f$), yielding
\begin{equation}
F_{\mathrm{cat}}=[F_S;F_T]\in\mathbb{R}^{H_f\times W_f\times 2D_f}.
\end{equation}
The concatenated feature map $F_{\mathrm{cat}}$ is projected to the U-Net's transformer token dimension $D_v$ via an MLP, then flattened into a sequence of $N_c$ conditioning tokens $\mathbf{c}=\{c_1,\dots,c_{N_c}\}$ where $N_c=H_f\times W_f$.

\begin{figure*}[htbp]
  \centering
  \includegraphics[width=1.0\columnwidth,trim=1cm 5.5cm 1cm 5.5cm,clip]{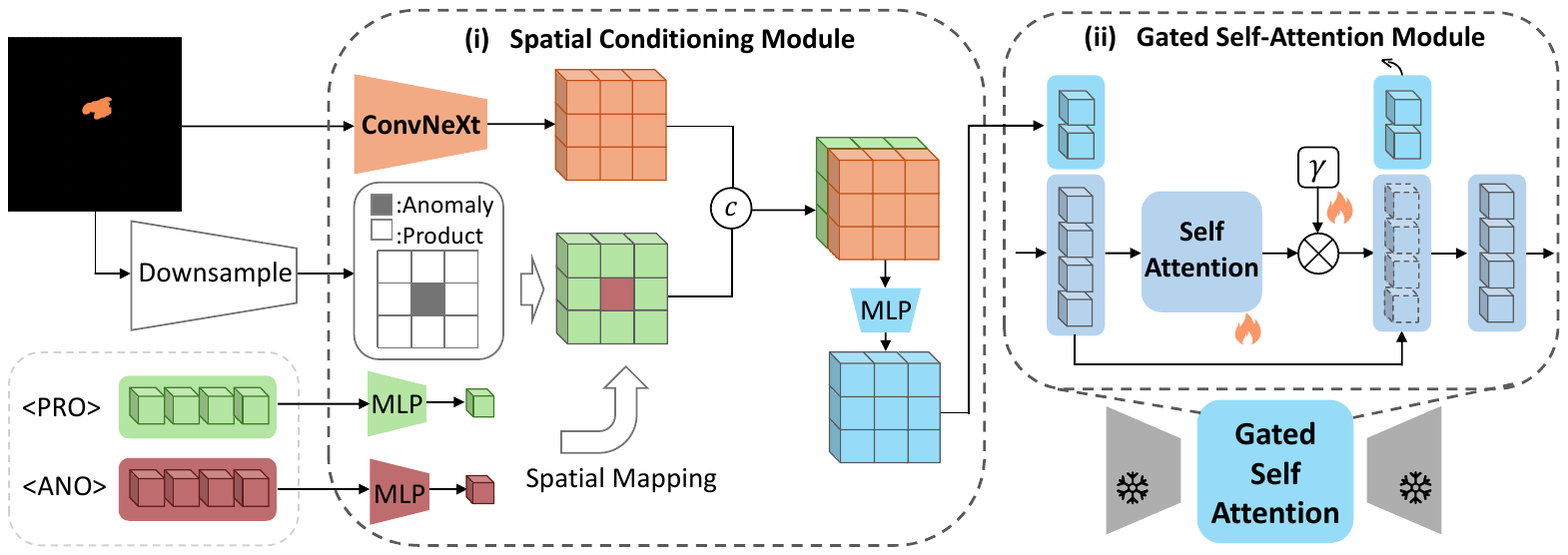}
  \caption{(i) Spatial Conditioning Module; (ii) Gated Self-Attention Module.}
  \label{fig:model}
\end{figure*}

\subsection{Gated Self-Attention Module}
\label{subsec:fuser}
\noindent Few-shot anomaly synthesis poses a trade-off: prior methods either freeze the U-Net and train only textual embeddings \cite{hu2023anomalydiffusion}, which limits adaptation capacity, or fine-tune the U-Net \cite{jin2025dual, dai2025SeaS}, which risks overwriting pretrained knowledge. To address this, we keep the U-Net frozen and introduce a \textbf{Gated Self-Attention Module}  that injects spatial–textual conditioning into transformer blocks in the U-Net via gated self-attention layers, enabling expressive adaptation without modifying original U-Net weights.
The conditioning tokens from the SCM are concatenated with the visual tokens to form \(\mathbf{h_c}=[\mathbf{v};\mathbf{c}]\), and self-attention is applied to the resulting sequence. The gated fusion updates the visual tokens as:
\begin{equation}
\mathbf{v} \leftarrow \mathbf{v} + \tanh(\gamma)\cdot \mathrm{Select}_v\big(\mathrm{SelfAttn}(\mathbf{h_c})\big),
\label{eq:gsa}
\end{equation}
where \(\gamma\) is a scalar parameter initialized to 0, and $\mathrm{Select}_v(\cdot)$ extracts the first $M$ tokens (corresponding to visual features) from the attention output while discarding the conditioning tokens. The gated attention is performed between the standard self-attention (\cref{eq:sa}) and cross-attention (\cref{eq:ca}) in each transformer block. 
Initialize \(\gamma=0\) to preserve the pretrained U-Net at the beginning of the training; the self-attention layers are augmented with LoRA \cite{hu2022lora} for low-rank updates, enabling stable few-shot adaptation.

\subsection{Normal-Guided Training and Synthesis}
\label{subsec:train}

\noindent
\textbf{Mixed Normal-anomalous Training.} While GSM enables effective few-shot adaptation, fine-tuning solely on limited anomalous exemplars risks severe overfitting, as their normal regions lack visual diversity. To mitigate this domain collapse, we propose a \textbf{mixed normal-anomalous training} (MNT) strategy. By incorporating abundant normal images, we regularize the model to preserve high-fidelity product priors while learning anomaly-specific patterns. 

Concretely, a unified model is trained per product across all anomaly types to enable feature sharing and reduce per-class data requirements. Normal images are conditioned with all-zero semantic maps $S=0$ and the prompt \texttt{"A photo of \{PRODUCT\}"}, teaching the model to reconstruct defect-free appearances. At each iteration, mini-batches are sampled from both normal and anomalous sets. This mixed sampling significantly improves anomaly fidelity and preserves global background consistency.

\noindent
\textbf{Normal-prior Denoising Initialization.}
To further bridge the domain gap between synthesized defects and real backgrounds, we introduce \textbf{normal-prior denoising initialization} (NDI) to leverage the visual priors of available normal images during the generation phase. Instead of initializing the reverse diffusion process from pure Gaussian noise $z_T \sim \mathcal{N}(0, I)$, we utilize the latent representation $z_0$ of a real normal image. 

Specifically, we inject noise into $z_0$ via the forward process to an intermediate timestep $t' < T$, yielding a partially noised latent $z_{t'}$. The reverse process then starts from $z_{t'}$, performing $t'$ denoising steps conditioned on the target semantic map and prompt to yield the final anomalous image. This initialization strategy offers two key advantages: (i) it yields higher generation quality by seamlessly integrating anomalies into realistic, highly-detailed backgrounds, and (ii) it significantly accelerates inference by reducing the required denoising steps from $T$ to $t'$.

\subsection{Mask Generation}
\label{subsec:mask_gen}

\noindent Spatial control requires diverse anomaly masks as input, yet real datasets typically provide very few masks per anomaly type. This scarcity limits the spatial variability of synthesized defects, reducing the effectiveness of downstream anomaly detectors trained on generated data. To ensure a sufficient and varied supply of masks for our pipeline, we adopt a standard approach by learning the underlying mask distribution via Textual Inversion~\cite{gal2022image, hu2023anomalydiffusion}. 

Specifically, we learn a dedicated token embedding, $e_m$, that encapsulates the concept of an "anomaly mask." For each anomaly type, this embedding is optimized by minimizing the standard diffusion loss with respect to the available real masks:
\begin{equation}
\label{eq:mask_ti}
e^*_m = \arg \min_{e_m} \mathbb{E}_{m\sim p_{\text{masks}},\,\epsilon\sim\mathcal{N}(0,I),\,t}
\left\| \epsilon - \epsilon_\theta(z_t, t, e_m) \right\|_2^2,
\end{equation}
where $z_t$ is the noised latent of a real mask $m$. Once optimized, this embedding $e_m^*$ can be used as a conditional text prompt to generate novel masks that preserve the shape and size characteristics of the original distribution while introducing spatial variations.

\section{Experiments}
\label{sec:exp}

\subsection{Experiment Settings}
\noindent\textbf{Datasets.}
Experiments are conducted on MVTec AD \cite{8954181} and VisA \cite{zou2022spot} datasets.
In \cref{subsec:qua,subsec:AGforAI}, following the protocol of SeaS \cite{dai2025SeaS}, 60 normal images per product and one-third of anomalous images per anomaly type are used for training, with the remaining two-thirds reserved for testing. In \cref{subsec:abla}, following the setup of SeaS \cite{dai2025SeaS}, which shows that using more than 2 images per anomaly type can induce train–test overlap (the minimum number of abnormal training images is 2 in previous setting), we conduct experiments under the 1-shot and 2-shot settings. 

\noindent\textbf{Implementation details.}
Following AnomalyDiffusion \cite{hu2023anomalydiffusion}, 1,000 anomaly image–mask pairs per anomaly type are generated to train inspection models. A single generative model is trained per product to cover all anomaly types. Additional experimental details are provided in the supplementary material. 

\noindent\textbf{Metrics.}
For generation quality, Inception Score (IS) and intra-cluster pairwise LPIPS (IC-LPIPS) are reported. 
For anomaly segmentation, Area Under the ROC curve (AUROC), Average Precision (AP), and \(F_1\)-max are used to evaluate image-level and pixel-level performance on MVTec and VisA; for instance-level anomaly detection, mean Average Precision (mAP) is reported. 

\noindent \textbf{Baselines.}
Our method is compared with state-of-the-art anomaly-synthesis approaches: DFMGAN \cite{Duan2023DFMGAN}, AnomalyDiffusion \cite{hu2023anomalydiffusion}, DualAnoDiff \cite{jin2025dual}, and SeaS \cite{dai2025SeaS}. GroundingAnomaly is also compared against recent anomaly detection methods: DRÆM \cite{Zavrtanik_2021_ICCV}, GLASS \cite{chen2024unified}, WinCLIP+\cite{jeong2023winclip}, PromptAD\cite{li2024promptad}, ReMP-AD\cite{ma2025remp}, PatchCore \cite{roth2022towards}, ViTAD \cite{zhang2023exploring}, MambaAD \cite{he2024mambaad}, INP-Former \cite{luo2025INP-Former} and Dinomaly2 \cite{guo2025one}. 

\subsection{Anomaly Generation Quality}
\label{subsec:qua}
\begin{figure}[ht]
    \centering
    \includegraphics[width=1\columnwidth, trim=2.1cm 2.8cm 2.1cm 2.8cm,clip]{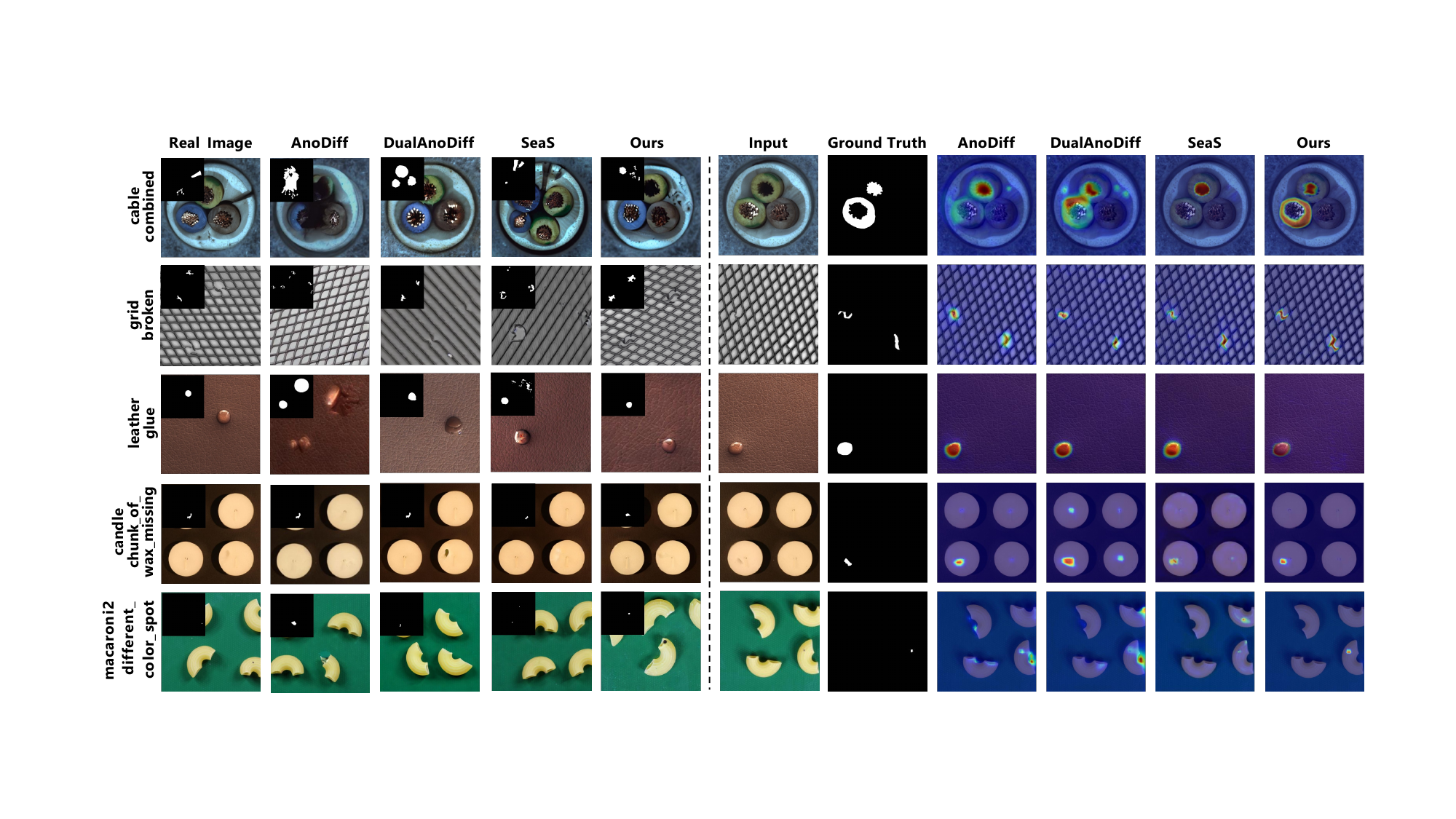}
    \caption{\textbf{Left:} Comparison on the generation results on MVTec AD and VisA. Cable, grid and leather belong to MVTec AD and candle and macaroni2 belong to VisA. Our method generates high-quality anomaly images that are accurately grounded with masks. \textbf{Right:} Visualization of U-Net segmentation results on MVTec AD and VisA. The detection model trained on GroundingAnomaly-generated data exhibits more robust performance.}
    \label{fig:result}
\end{figure}

\noindent GroundingAnomaly is compared with baselines on anomalous image generation quality (IS) and diversity (IC-LPIPS) on MVTec AD and VisA (see \cref{tab:IS_ICLPIPS}). The results demonstrate that our method achieves the highest quality and diversity among the evaluated approaches.
Generated examples for MVTec AD and VisA are shown in \cref{fig:result}. Compared to prior methods, images produced by GroundingAnomaly exhibit significantly improved visual fidelity with more controlled backgrounds and tighter alignment between anomaly masks and defect regions. 

\begin{table}[b]
  \centering
  \scriptsize
  \setlength{\tabcolsep}{4pt}
  \renewcommand{\arraystretch}{0.95}
  \caption{Comparison on IS and IC-LPIPS on MVTec AD and VisA. Bold indicates the best performance.}
  \resizebox{\columnwidth}{!}{%
  \begin{tabular}{l|cc|cc|cc|cc|cc}
    \toprule
    Dataset
      & \multicolumn{2}{c|}{DFMGAN\cite{Duan2023DFMGAN}}
      & \multicolumn{2}{c|}{AnoDiff\cite{hu2023anomalydiffusion}}
      & \multicolumn{2}{c|}{DualAnoDiff\cite{jin2025dual}}
      & \multicolumn{2}{c|}{SeaS\cite{dai2025SeaS}}
      & \multicolumn{2}{c}{\textbf{Ours}} \\
    & IS  & IC-L 
    & IS  & IC-L 
    & IS  & IC-L 
    & IS  & IC-L  
    & IS  & IC-L  \\
    \midrule
    MVTec AD
      & 1.72 & 0.20 
      & 1.80 & 0.32 
      & 1.91 & 0.37 
      & 1.95 & 0.34 
      & \textbf{1.99} & \textbf{0.41}  \\
    VisA 
      & 1.25 & 0.26 
      & 1.26 & 0.25 
      & 1.25 & 0.25 
      & 1.27 & 0.26 
      & \textbf{1.29} & \textbf{0.31}  \\
    \bottomrule
  \end{tabular}%
  }
  
  \label{tab:IS_ICLPIPS}
\end{table}

\subsection{Anomaly Generation for Anomaly Inspection}
\label{subsec:AGforAI}

\noindent \textbf{Anomaly image generation for anomaly detection and segmentation. }
To evaluate downstream anomaly detection and segmentation performance, a simple U-Net \cite{ronneberger2015u} is trained per product using the image–mask pairs synthesized by each method. Pixel-level segmentation outputs are converted to image-level anomaly confidence scores via global average pooling. 
MVTec AD results are reported in \cref{tab:unet_mvtec}. GroundingAnomaly substantially improves segmentation performance, notably on the \textit{grid} and \textit{screw} categories with AP gains of 5.9\% and 2.7\%, respectively, and achieves the highest AP (85.9\%) and \(F_1\)-max (81.8\%) on MVTec AD. VisA results are presented in \cref{tab:unet_VisA}. Our method yields AP improvements of 16.6\% and 15.2\% on the \textit{capsules} and \textit{macaroni2} categories, respectively, attaining the highest overall AP (67.2\%) on VisA and outperforming the second-ranked DualAnoDiff by 6.3\% AP.

Visualized results are presented in \cref{fig:result}, which shows that segmentation U-Nets trained on image–mask pairs generated by GroundingAnomaly exhibit better alignment with the ground truth with fewer false positives and higher prediction confidence. 
GroundingAnomaly is also compared with state-of-the-art anomaly inspection methods in \cref{tab:compare}, which shows that simple U-Nets trained on GroundingAnomaly-generated data outperform these baselines by 11.8\% AP on MVTec AD and 12.5\% AP on VisA, while achieving competitive AUROC.

\begin{table*}[t]
  \centering
  \scriptsize
  \setlength{\tabcolsep}{4pt}
  \renewcommand{\arraystretch}{0.95}
  \caption{Comparison on pixel-level segmentation and image-level detection on MVTec AD dataset by training U-Nets on the generated data from DRÆM, DFMGAN, AnomalyDiffusion, DualAnoDiff, SeaS and our proposed GroundingAnomaly. }
  \resizebox{\textwidth}{!}{
  \begin{tabular}{l|cccc|cccc|cccc|cccc|cccc|cccc}
    \toprule
    Category
    & \multicolumn{4}{c|}{DRÆM \cite{Zavrtanik_2021_ICCV}}
    & \multicolumn{4}{c|}{DFMGAN \cite{Duan2023DFMGAN}}
    & \multicolumn{4}{c|}{AnomalyDiffusion \cite{hu2023anomalydiffusion}}
    & \multicolumn{4}{c|}{DualAnoDiff \cite{jin2025dual}}
    & \multicolumn{4}{c|}{SeaS \cite{dai2025SeaS}}
    & \multicolumn{4}{c}{Ours} \\
    & AUC-P & AP-P & F1-P & AP-I
    & AUC-P & AP-P & F1-P & AP-I
    & AUC-P & AP-P & F1-P & AP-I
    & AUC-P & AP-P & F1-P & AP-I
    & AUC-P & AP-P & F1-P & AP-I
    & AUC-P & AP-P & F1-P & AP-I \\
    \midrule
    bottle
      & 96.7 & 80.2 & 74.0 & 99.8
      & 98.9 & 90.2 & 83.9 & 99.8
      & 99.4 & 94.1 & 87.3 & 99.9
      & 99.5 & 93.4 & 85.7 & \textbf{100.0}
      & \textbf{99.7} & \textbf{95.5} & 87.9 & \textbf{100.0}
      & 99.6 & 94.1 & \textbf{88.1} & \textbf{100.0} \\
    cable
      & 80.3 & 21.8 & 28.3 & 83.2
      & 97.2 & 81.0 & 75.4 & 97.8
      & \textbf{99.2} & \textbf{90.8} & \textbf{83.5} & \textbf{100.0}
      & 97.5 & 82.6 & 76.9 & 98.3
      & 94.5 & 77.8 & 75.0 & 98.5
      & 97.9 & 81.0 & 78.2 & 99.5 \\
    capsule
      & 76.2 & 25.5 & 32.1 & 98.7
      & 79.2 & 26.0 & 35.0 & 98.5
      & 98.8 & 57.2 & 59.8 & \textbf{99.9}
      & \textbf{99.5} & \textbf{73.2} & \textbf{67.0} & 99.2
      & 95.7 & 51.7 & 52.4 & 99.6
      & 98.3 & 65.5 & 59.4 & 98.8 \\
    carpet
      & 92.6 & 43.0 & 41.9 & 98.7
      & 90.6 & 33.4 & 38.1 & 98.5
      & 98.6 & 81.2 & 74.6 & 98.8
      & 99.4 & 89.1 & 80.2 & \textbf{99.9}
      & \textbf{99.5} & 84.5 & 76.5 & 99.1
      & \textbf{99.5} & \textbf{89.3} & \textbf{86.3} & 99.6 \\
    grid
      & 99.1 & 59.3 & 58.7 & 99.9
      & 75.2 & 14.3 & 20.5 & 90.4
      & 98.3 & 52.9 & 54.6 & 99.5
      & 98.5 & 57.2 & 54.9 & 99.7
      & 99.6 & 67.7 & 62.1 & 99.9
      & \textbf{99.7} & \textbf{73.6} & \textbf{69.9} & \textbf{100.0} \\
    hazelnut
      & 98.8 & 73.6 & 68.5 & \textbf{100.0}
      & 99.7 & 95.2 & 89.5 & \textbf{100.0}
      & \textbf{99.8} & 96.5 & 90.6 & 99.9
      & \textbf{99.8} & \textbf{97.7} & \textbf{92.8} & \textbf{100.0}
      & 99.7 & 91.4 & 84.3 & \textbf{100.0}
      & 99.6 & 94.4 & 89.4 & \textbf{100.0} \\
    leather
      & 98.5 & 67.6 & 65.0 & \textbf{100.0}
      & 98.5 & 68.7 & 66.7 & \textbf{100.0}
      & 99.8 & 79.6 & 71.0 & \textbf{100.0}
      & \textbf{99.9} & 88.8 & 78.8 & \textbf{100.0}
      & 99.2 & 69.1 & 64.9 & 99.9
      & \textbf{99.9} & \textbf{89.2} & \textbf{88.1} & \textbf{100.0} \\
    metal nut
      & 96.9 & 84.2 & 74.5 & 99.6
      & 99.3 & 98.1 & 94.5 & 99.8
      & 99.8 & 98.7 & 94.0 & \textbf{100.0}
      & 99.6 & 98.0 & 93.0 & 99.9
      & 99.8 & 98.8 & 94.3 & \textbf{100.0}
      & \textbf{99.9} & \textbf{99.3} & \textbf{95.7} & \textbf{100.0} \\
    pill
      & 95.8 & 45.3 & 53.0 & 98.9
      & 81.2 & 67.8 & 72.6 & 91.7
      & \textbf{99.8} & \textbf{97.0} & \textbf{90.8} & 99.6
      & 99.6 & 95.8 & 89.2 & 99.0
      & 99.6 & 90.2 & 82.7 & \textbf{99.8}
      & 99.2 & 95.0 & 89.7 & 98.8 \\
    screw
      & 91.0 & 30.1 & 35.7 & 96.3
      & 58.8 & 2.2 & 5.3 & 64.7
      & 97.0 & 51.8 & 50.9 & \textbf{97.9}
      & 98.1 & 57.1 & \textbf{56.1}& 95.0
      & \textbf{98.3} & 55.2 & 54.7 & 97.6
      & 97.8 & \textbf{59.8} & 55.3 & 91.9 \\
    tile
      & 98.5 & 93.2 & 87.8 & \textbf{100.0}
      & 99.5 & 97.1 & 91.6 & \textbf{100.0}
      & 99.2 & 93.9 & 86.2 & \textbf{100.0}
      & 99.7 & 97.1 & 91.0 & \textbf{100.0}
      & \textbf{99.8} & \textbf{97.2} & 91.7 & \textbf{100.0}
      & 99.5 & 96.2 & \textbf{93.1} & \textbf{100.0} \\
    toothbrush
      & 93.8 & 29.5 & 28.4 & 99.8
      & 96.4 & 75.9 & 72.6 & \textbf{100.0}
      & 99.2 & 76.5 & 73.4 & \textbf{100.0}
      & 98.2 & 68.3 & 68.6 & 99.7
      & 96.1 & 57.5 & 58.8 & 95.9
      & \textbf{99.3} & \textbf{79.2} & \textbf{75.7} & \textbf{100.0} \\
    transistor
      & 76.5 & 31.7 & 24.2 & 80.5
      & 96.2 & 81.2 & 77.0 & 92.5
      & 99.3 & \textbf{92.6} & 85.7 & \textbf{100.0}
      & 98.0 & 86.7 & 79.6 & 93.7
      & 96.9 & 80.5 & 77.6 & \textbf{100.0}
      & \textbf{99.6} & 92.3 & \textbf{88.7} & \textbf{100.0} \\
    wood
      & 98.8 & 87.8 & 80.9 & \textbf{100.0}
      & 95.3 & 70.7 & 65.8 & 99.4
      & 98.9 & 84.6 & 74.5 & 99.4
      & 99.4 & \textbf{91.6} & 83.8 & 99.9
      & 99.4 & 87.2 & 79.2 & \textbf{100.0}
      & \textbf{99.5} & 90.7 & \textbf{86.7} & \textbf{100.0} \\
    zipper
      & 93.4 & 65.4 & 64.7 & \textbf{100.0}
      & 92.9 & 65.6 & 64.9 & 99.9
      & 99.4 & 86.0 & 79.2 & \textbf{100.0}
      & \textbf{99.6} & \textbf{90.7} & \textbf{82.7} & \textbf{100.0}
      & 99.3 & 85.8 & 79.0 & \textbf{100.0}
      & \textbf{99.6} & 88.7 & 82.2 & \textbf{100.0} \\
    \midrule
    Average
      & 92.5 & 55.9 & 54.5 & 97.0
      & 90.6 & 64.5 & 63.6 & 95.5
      & 99.1 & 82.2 & 77.1 & \textbf{99.7}
      & 99.1 & 84.5 & 78.7 & 99.0
      & 98.5 & 79.3 & 74.7 & 99.4
      & \textbf{99.3} & \textbf{85.9} & \textbf{81.8} & 99.2 \\
    \bottomrule
  \end{tabular}
  }
  \label{tab:unet_mvtec}
\end{table*}

\begin{table*}[htbp]
  \centering
  \scriptsize
  \setlength{\tabcolsep}{4pt}
  \renewcommand{\arraystretch}{0.95}
    \caption{Comparison on pixel-level segmentation and image-level detection on VisA dataset by training U-Nets on the generated data from DRÆM, DFMGAN, AnomalyDiffusion, DualAnoDiff, SeaS and our proposed GroundingAnomaly. }
  \resizebox{\textwidth}{!}{
  \begin{tabular}{l|
    c c c c|
    c c c c|
    c c c c|
    c c c c|
    c c c c|
    c c c c}
    \toprule
    Category
      & \multicolumn{4}{c|}{DRÆM \cite{Zavrtanik_2021_ICCV}}
      & \multicolumn{4}{c|}{DFMGAN \cite{Duan2023DFMGAN}}
      & \multicolumn{4}{c|}{AnomalyDiffusion \cite{hu2023anomalydiffusion}}
      & \multicolumn{4}{c|}{DualAnoDiff \cite{jin2025dual}}
      & \multicolumn{4}{c|}{SeaS \cite{dai2025SeaS}}
      & \multicolumn{4}{c}{Ours} \\
      & AUC-P & AP-P & F1-P & AP-I
      & AUC-P & AP-P & F1-P & AP-I
      & AUC-P & AP-P & F1-P & AP-I
      & AUC-P & AP-P & F1-P & AP-I
      & AUC-P & AP-P & F1-P & AP-I
      & AUC-P & AP-P & F1-P & AP-I \\
    \midrule
    candle
      & 86.2 & 18.7 & 23.7 & 77.5
      & 87.1 & 24.7 & 31.4 & 85.9
      & 89.1 & 28.7 & 33.9 & 87.2
      & 93.7 & 48.1 & 46.5 & 89.0
      & 89.4 & 45.1 & 44.1 & 87.8
      & \textbf{98.9} & \textbf{61.2} & \textbf{58.3} & \textbf{94.4} \\
    capsules
      & 90.7 & 50.4 & 53.8 & 85.1
      & 91.3 & 55.0 & 58.6 & 86.5
      & 94.1 & 60.5 & 62.3 & 90.7
      & 95.8 & 61.8 & 64.5 & 92.0
      & 97.2 & 61.2 & 64.2 & 88.7
      & \textbf{99.3} & \textbf{78.4} & \textbf{72.1} & \textbf{92.3} \\
    cashew
      & 89.0 & 45.2 & 47.1 & 91.1
      & 92.2 & 65.7 & 65.1 & 92.3
      & 95.1 & 82.1 & 80.9 & 95.0
      & 96.5 & 83.1 & \textbf{87.1} & \textbf{96.1}
      & 97.0 & 80.9 & 81.3 & 93.3
      & \textbf{99.3} & \textbf{87.1} & 84.2 & 95.3 \\
    chewinggum
      & 94.3 & 59.7 & 65.0 & 97.0
      & 96.0 & 67.4 & 63.0 & 96.7
      & 97.3 & 80.5 & 75.3 & 95.5
      & 98.8 & 80.3 & 77.1 & 98.8
      & 98.5 & 81.2 & 74.8 & 99.3
      & \textbf{99.5} & \textbf{84.6} & \textbf{78.2} & \textbf{99.4} \\
    fryum
      & 88.2 & 41.3 & 34.1 & 81.9
      & 90.4 & 45.7 & 43.2 & 89.3
      & 93.4 & 57.1 & 52.9 & 92.1
      & 94.8 & 62.9 & 56.8 & 94.4
      & \textbf{95.8} & \textbf{63.8} & \textbf{59.0} & 93.6
      & 91.5 & 55.1 & 51.3 & \textbf{96.4} \\
    macaroni1
      & 94.3 & 38.9 & 34.3 & 88.0
      & 96.2 & 42.1 & 46.1 & 93.3
      & 97.2 & 48.7 & 51.3 & 94.8
      & 98.7 & 53.9 & 53.7 & 98.0
      & 98.1 & 47.7 & 50.8 & 93.5
      & \textbf{99.8} & \textbf{65.4} & \textbf{59.4} & \textbf{99.6} \\
    macaroni2
      & 85.7 & 30.1 & 32.9 & 80.4
      & 89.3 & 32.0 & 35.7 & 81.4
      & 94.1 & 35.2 & 41.1 & 85.3
      & 97.1 & 39.3 & 45.7 & 88.0
      & 96.7 & 39.1 & 44.4 & 87.7
      & \textbf{99.3} & \textbf{54.5} & \textbf{50.9} & \textbf{91.9} \\
    pcb1
      & 91.1 & 72.2 & 71.4 & 95.4
      & 93.0 & 79.4 & 74.3 & 97.1
      & 96.3 & 80.7 & 77.1 & 97.9
      & 97.1 & 81.4 & \textbf{78.0} & \textbf{98.9}
      & \textbf{97.5} & \textbf{81.9} & 77.9 & 98.8
      & 96.7 & 81.8 & 75.7 & 97.8 \\
    pcb2
      & 88.2 & 27.4 & 21.7 & 94.9
      & 90.4 & 32.7 & 34.3 & 95.1
      & 92.9 & 42.5 & 43.7 & 96.9
      & 95.0 & 48.0 & 47.2 & \textbf{97.3}
      & 94.7 & 47.7 & 46.5 & 97.2
      & \textbf{97.5} & \textbf{54.0} & \textbf{55.4} & 96.1 \\
    pcb3
      & 93.3 & 38.1 & 39.5 & 92.1
      & 93.8 & 50.1 & 36.7 & 94.3
      & 94.2 & 51.9 & 47.3 & 94.9
      & 96.7 & 53.2 & 55.9 & \textbf{95.8}
      & 95.9 & 48.3 & 50.2 & 95.0
      & \textbf{97.7} & \textbf{61.6} & \textbf{61.0} & \textbf{95.8} \\
    pcb4
      & 90.1 & 48.3 & 43.1 & 92.1
      & 93.3 & 50.4 & 44.9 & 93.7
      & 95.3 & 51.7 & 53.2 & 94.9
      & 98.7 & \textbf{58.3} & \textbf{58.1} & 98.0
      & 97.8 & 57.2 & 52.8 & \textbf{98.7}
      & \textbf{99.4} & 45.1 & 45.9 & 97.8 \\
    pipe fryum
      & 84.4 & 41.7 & 47.1 & 82.1
      & 88.3 & 44.3 & 48.3 & 85.7
      & 92.2 & 55.4 & 62.4 & 86.5
      & 96.4 & 60.3 & 65.7 & 90.1
      & 98.2 & 64.0 & 66.5 & 88.3
      & \textbf{99.5} & \textbf{77.4} & \textbf{68.7} & \textbf{95.7} \\
    \midrule
    Average
      & 89.6 & 42.7 & 42.8 & 88.1
      & 91.8 & 49.1 & 48.5 & 90.9
      & 94.3 & 56.3 & 56.8 & 92.6
      & 96.6 & 60.9 & 61.4 & 94.7
      & 96.4 & 59.8 & 59.3 & 93.5
      & \textbf{98.2} & \textbf{67.2} & \textbf{63.5} & \textbf{96.0} \\
    \bottomrule
  \end{tabular}
  }
  \label{tab:unet_VisA}
\end{table*}

\noindent \textbf{Anomaly image generation for instance-level anomaly detection.} 
To assess instance-level detection utility, object detectors (Faster R-CNN \cite{ren2015faster}, DETR \cite{carion2020end} and YOLOv5 \cite{glenn_jocher_2020_4154370} ) are trained separately for each product on the synthesized image–mask pairs; instance bounding boxes are derived from tight bounding boxes of masks. Mixed-type (category \textit{combined}) anomalies are excluded to avoid label ambiguity. The averaged mAP are reported in \cref{tab:instance}, showing average mAP improvements of 1.52\% (MVTec AD) and 2.34\% (VisA). Our method achieves superior performance by generating anomalies with more class-aligned appearances.

\begin{table}[htbp]
  \centering
  \scriptsize
  \setlength{\tabcolsep}{4pt}
  \renewcommand{\arraystretch}{0.95}
  \begin{minipage}[t]{0.49\textwidth}
    \centering
    \captionof{table}{Comparison on pixel-level segmentation and image-level detection with anomaly inspection methods.}
    \label{tab:compare}
    \resizebox{\linewidth}{!}{%
      \begin{tabular}{l|
    >{\centering\arraybackslash}p{1cm}
    >{\centering\arraybackslash}p{1cm}
    >{\centering\arraybackslash}p{1cm}
    >{\centering\arraybackslash}p{1cm}
    |>{\centering\arraybackslash}p{1cm}
    >{\centering\arraybackslash}p{1cm}
    >{\centering\arraybackslash}p{1cm}
    >{\centering\arraybackslash}p{1cm}}%
    \toprule
    Model
      & \multicolumn{4}{c|}{\textbf{MVTec AD}}
      & \multicolumn{4}{c}{\textbf{VisA}} \\
      & AUC-P & AP-P & F1-P & AP-I
      & AUC-P & AP-P & F1-P & AP-I \\
    \midrule
    \multicolumn{9}{l}{\textbf{Unsupervised}} \\
    \midrule
    PatchCore\cite{roth2022towards}
      & 98.4  & 56.1 & 58.9  & 98.6
      & 98.4  & 48.6 & 49.7  & 94.8 \\
    ViTAD\cite{zhang2023exploring}
      & 97.7  & 55.3  & 58.7  & 98.3
      & 98.2  & 36.6  & 41.1  & 90.5 \\
    MambaAD\cite{he2024mambaad}
      & 97.7  & 56.3  & 59.2  & 93.1
      & 98.5  & 39.4  & 44.0  & 94.3 \\
    WinCLIP+\cite{jeong2023winclip}
      & 96.2  & 62.7  & 60.1  & 98.8
      & 97.4  & 50.2  & 47.9  & 89.8 \\
    PromptAD\cite{li2024promptad}
      & 96.3  & 63.4  & 60.3  & 98.9
      & 97.5  & 50.3  & 47.6  & 91.4 \\
    ReMP-AD\cite{ma2025remp}
      & 96.7  & 64.8  & 61.2  & 99.2
      & 97.9  & 51.2  & 49.1  & 94.8 \\
    INP-Former\cite{luo2025INP-Former}
      & 98.5  & 71.0  & 69.7  & 99.9
      & 98.9  & 51.2  & 54.7  & 99.0 \\
    Dinomaly2\cite{guo2025one}
      & 98.6  & 70.3  & 69.9  & \textbf{100.0}
      & \textbf{99.2} & 54.7  & 57.1  & \textbf{99.3} \\
    \midrule
    \multicolumn{9}{l}{\textbf{Synthesize-based}} \\
    \midrule
    DRÆM\cite{Zavrtanik_2021_ICCV}
      &97.9 &67.9 &66.1  & 98.0
      & 92.9 &17.2 &23.0  & 86.3 \\
    GLASS\cite{chen2024unified}
      & 99.3  & 74.1  & 70.4  & 99.9
      & 98.5 & 45.6 & 48.4  & 97.7 \\
    \midrule
    \textbf{Ours}+\textbf{U-Net}
      & \textbf{99.3} & \textbf{85.9} & \textbf{81.8} & 99.2
      & 98.2  & \textbf{67.2} & \textbf{63.5} & 96.0 \\
    \bottomrule
  \end{tabular}%
    }
  \end{minipage}%
  \hfill
  \begin{minipage}[t]{0.49\textwidth}
    \centering
    \captionof{table}{Comparison on trained supervised instance-level anomaly detection models on MVTec AD and VisA. }
    \label{tab:instance}
    \resizebox{\linewidth}{!}{%
      \begin{tabular}{c|
        >{\centering\arraybackslash}p{1.5cm}
        >{\centering\arraybackslash}p{1.5cm}
        >{\centering\arraybackslash}p{1.5cm}
        >{\centering\arraybackslash}p{1.5cm}
        >{\centering\arraybackslash}p{1.5cm}}
        \toprule
        Model       & DFMGAN & AnoDiff & DualAnoDiff & SeaS & \textbf{Ours} \\
        & avg mAP & avg mAP  & avg mAP  & avg mAP & avg mAP \\
        \midrule
        \multicolumn{6}{l}{\textbf{MVTec}} \\
        \midrule
        Faster R-CNN \cite{ren2015faster}          & 35.44 & 37.13 & 40.17 & 40.44 & \textbf{41.96} \\
        DETR \cite{carion2020end}                 & 37.88 & 38.91 & 40.79 & 42.81 & \textbf{43.17} \\
        YOLOv5 \cite{glenn_jocher_2020_4154370}   & 41.51 & 44.35 & 47.73 & 46.71 & \textbf{49.40} \\
        \midrule
        Average     & 38.28 & 40.13 & 42.90 & 43.32 & \textbf{44.84} \\
        \midrule
        \addlinespace[6pt]
        \multicolumn{6}{l}{\textbf{VisA}} \\
        \midrule
        Faster R-CNN \cite{ren2015faster}          & 28.76 & 34.19 & \textbf{37.73} & 36.77 & 37.41 \\
        DETR \cite{carion2020end}                 & 30.17 & 32.07 & 35.16 & 33.95 & \textbf{38.77} \\
        YOLOv5 \cite{glenn_jocher_2020_4154370}   & 29.88 & 33.05 & 36.83 & 37.11 & \textbf{40.54} \\
        \midrule
        Average       & 29.60 & 33.10 & 36.57 & 35.94 & \textbf{38.91}\\
        \bottomrule
      \end{tabular}%
    }
  \end{minipage}
\end{table}

\subsection{Ablation Study}
\label{subsec:abla}

\noindent \textbf{Ablation on model components.} We evaluate each component via ablations: (i) without Disentangled Token Learning (fixed text tokens, w/o DTL); (ii) without Spatial–Textual Feature Fusion (w/o SFF); (iii) without GSM, injecting conditioning into U-Net cross-attention (w/o GSM); (iv) trained only on anomalous images (w/o MNT); (v) initialize generation from random noise (w/o NDI); and (vi) the full GroundingAnomaly. We use these models to generate 1,000 anomaly image-mask pairs per anomaly type and train a U-Net per product. The results are provided in \cref{tab:ablation}, which shows that removing any proposed module degrades generation quality and anomaly inspection performance. Note that a high IC-LPIPS combined with a low IS (w/o GSM on MVTec) indicates poor fidelity and chaotic generation rather than diversity.

\noindent \textbf{Ablation on few-shot generation.} 
In previous experiments, we follow the setting in AnomalyDiffusion \cite{hu2023anomalydiffusion} and employed 1/3 of anomaly data for comparison. However, most industrial applications require synthesizing anomalies from only a few exemplars. Here we analyze GroundingAnomaly's extreme few-shot capability. Following the setup of SeaS \cite{dai2025SeaS}, which shows that using more than 2 images per anomaly type can introduce train–test overlap (the minimum number of abnormal training images is 2 in previous setting), we conduct experiments under the 1-shot and 2-shot settings, results are provided in \cref{tab:few}, more qualitative results can be found in the supplementary material. In the 1-shot setting, GroundingAnomaly can generate images of satisfactory quality, but downstream detection performance is severely degraded by limited mask diversity, performance improves modestly in the 2-shot setting.

\subsection{Analysis}

\begin{figure}[htbp] 
\centering 
\includegraphics[width=1.0\columnwidth, trim=1.2cm 2.2cm 1.2cm 2.2cm,clip]{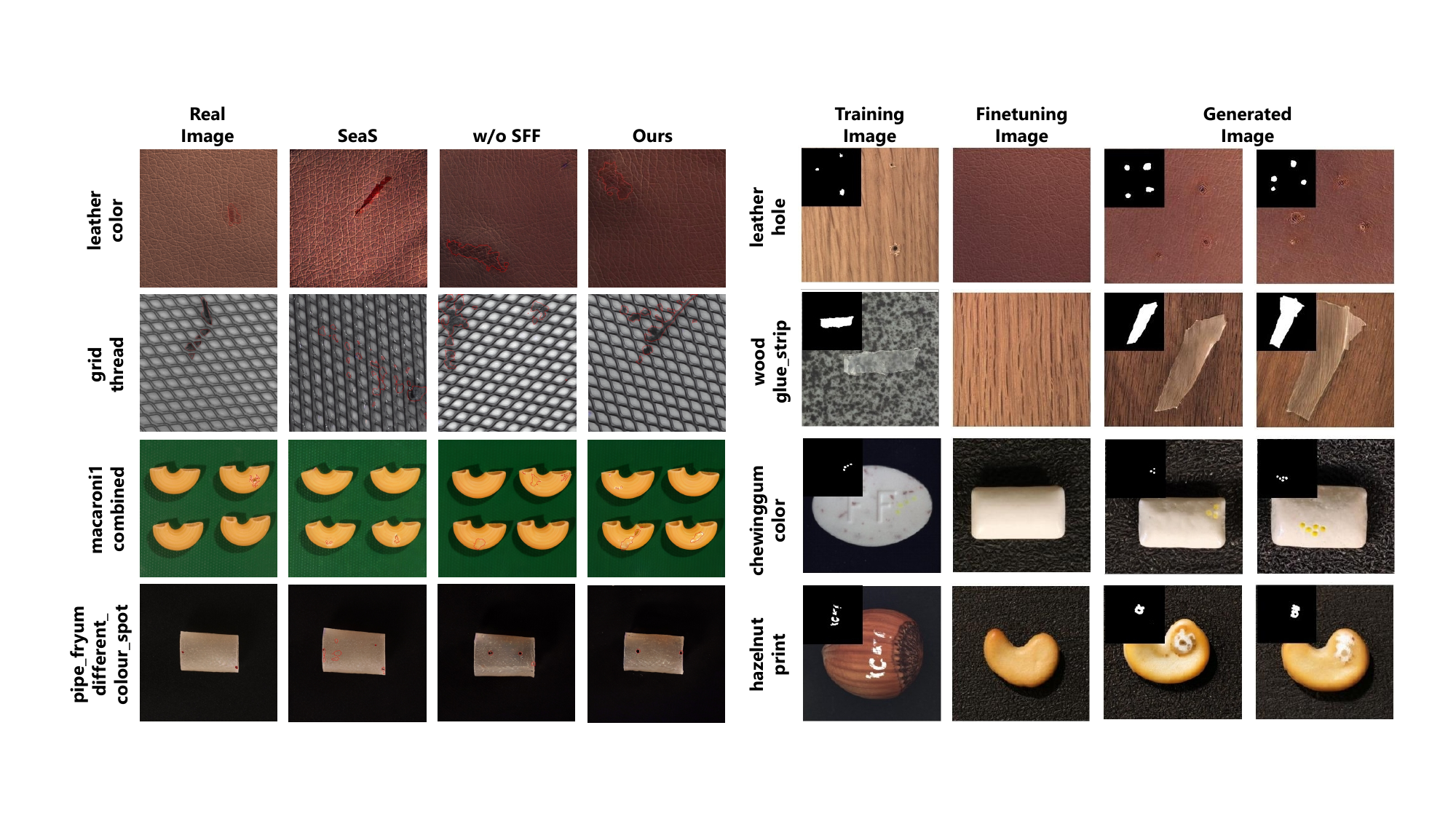} 
\caption{\textbf{Left:} Qualitative analysis of SFF. The red outlines denote the ground-truth mask. While SeaS \cite{dai2025SeaS} suffers from spatial misalignment, our model without SFF (w/o SFF) achieves accurate bounding. Our full model (Ours) leverages SFF to achieve both precise spatial grounding and high-fidelity anomaly synthesis. \textbf{Right:} Unseen anomaly generation results on MVTec AD and VisA. GroundingAnomaly is trained on anomalous images and finetuned on normal images from another product.} 
\label{fig:unseen} 
\end{figure}

\noindent \textbf{Analysis of Spatial-Textual Feature Fusion.}
While the quantitative ablation in \cref{tab:ablation} confirms the overall effectiveness of SFF, it does not explicitly reveal the source of this improvement. Qualitative analysis (\cref{fig:unseen}) provides the answer: SFF primarily enhances the visual realism of the synthesized anomalies rather than their spatial grounding. As illustrated, SeaS \cite{dai2025SeaS} struggles to align masks with the generated defects. Conversely, our model without SFF (w/o SFF) achieves accurate spatial bounding but often yields artificial textures and blunt transitions. By utilizing spatially aligned tokens to learn both anomaly and product features, SFF significantly enhances the overall generation quality and the realism of the synthesized defects.

\noindent \textbf{Unseen anomaly generation.} Beyond reconstructing known defects, the proposed GroundingAnomaly can synthesize novel anomalies on unseen products through brief fine-tuning on normal samples. \cref{fig:unseen} illustrates this cross-domain generalization: a model trained on \textit{wood} successfully projects learned anomaly characteristics onto \textit{leather}, generating realistic defects that are entirely absent from the target dataset's real distribution. The generated results confirm that our approach deeply comprehends anomaly structures and can synthesize diverse, out-of-distribution defects across various datasets, rather than merely overfitting to training exemplars.

\begin{table}[htbp]
  \centering
  \scriptsize
  \setlength{\tabcolsep}{4pt}
  \renewcommand{\arraystretch}{0.95}
  \begin{minipage}[t]{0.49\textwidth}
    \centering
    \captionof{table}{Ablation study results. }
    \label{tab:ablation}
    \resizebox{\linewidth}{!}{%
      \begin{tabular}{c|
        >{\centering\arraybackslash}p{1.35cm}
        >{\centering\arraybackslash}p{1.35cm}
        >{\centering\arraybackslash}p{1.35cm}
        >{\centering\arraybackslash}p{1.35cm}
        >{\centering\arraybackslash}p{1.35cm}}
        \toprule
        Model       & IS & IC-L & AUC-P & AP-P & AP-I\\
        \midrule
        \multicolumn{6}{l}{\textbf{MVTec AD}} \\
        \midrule
        w/o DTL       & 1.88 & 0.39 & 98.7 & 81.7 & 98.6 \\
        w/o SFF       & 1.72 & 0.37 & 97.5 & 77.3 & 97.7 \\
        w/o GSM       & 1.63 & \textbf{0.42} & 95.7 & 75.7 & 96.1\\
        w/o MNT       & 1.74 & 0.37 & 99.1 & 83.1 & 99.1\\
        w/o NDI       & 1.94 & 0.40 & 99.2 & 84.7 & 99.1 \\
        \midrule
        \textbf{Ours}       & \textbf{1.99} & 0.41 & \textbf{99.3} & \textbf{85.9} &\textbf{99.2} \\
        \midrule
        \multicolumn{6}{l}{\textbf{VisA}} \\
        \midrule
        w/o DTL       & 1.21 & 0.27 & 95.2 & 62.3 & 94.8 \\
        w/o SFF       & 1.13 & 0.28 & 93.9 & 59.1 &  93.7  \\
        w/o GSM       & 1.09 & 0.30 & 93.7 & 58.1 & 93.4 \\
        w/o MNT       & 1.17 & 0.27 & 97.3 & 62.9 & 95.3 \\
        w/o NDI       & 1.25 & 0.29 & 97.6 & 64.9 & 95.9 \\
        \midrule
        \textbf{Ours}       & \textbf{1.29} & \textbf{0.31} & \textbf{97.7} & \textbf{67.2} &\textbf{96.0} \\
        \bottomrule
      \end{tabular}
    }
  \end{minipage}
  \hfill
  \begin{minipage}[t]{0.49\textwidth}
    \centering
    \captionof{table}{Ablation on few-shot generation. }
    \label{tab:few}
    \resizebox{\linewidth}{!}{%
      \begin{tabular}{c|
    >{\centering\arraybackslash}p{0.9cm}
    >{\centering\arraybackslash}p{0.9cm}
    >{\centering\arraybackslash}p{0.9cm}
    >{\centering\arraybackslash}p{0.9cm}
    >{\centering\arraybackslash}p{0.9cm}}
    \toprule
    Model
      & IS & IC-L & AUC-P & AP-P & AP-I\\
    \midrule
    \multicolumn{6}{l}{\textbf{MVTec AD}} \\
    \midrule
    1-shot
      & 1.77 & 0.39 & 96.1 & 72.3 & 92.7 \\
    2-shot
      & 1.83 & 0.38 & 97.2 & 76.4 & 97.9 \\
      \midrule
    \textbf{Ours}
      & \textbf{1.99} & \textbf{0.41} & \textbf{99.3} & \textbf{85.9} &\textbf{99.2} \\
    \midrule
    \multicolumn{6}{l}{\textbf{VisA}} \\
    \midrule
    1-shot
      & 1.22 & 0.30 & 81.7 & 44.3 & 84.9 \\
    2-shot
      & 1.19 & 0.30 & 83.1 & 49.3 & 85.7  \\
      \midrule
    \textbf{Ours}
      & \textbf{1.29} & \textbf{0.31} & \textbf{97.7} & \textbf{67.2} &\textbf{96.0} \\
    \bottomrule
  \end{tabular}
    }
  \end{minipage}
\end{table}

\begin{wrapfigure}[15]{r}{0.45\columnwidth}
\centering
\includegraphics[width=0.45\columnwidth, trim=7.5cm 0.5cm 7.5cm 0.5cm,clip]{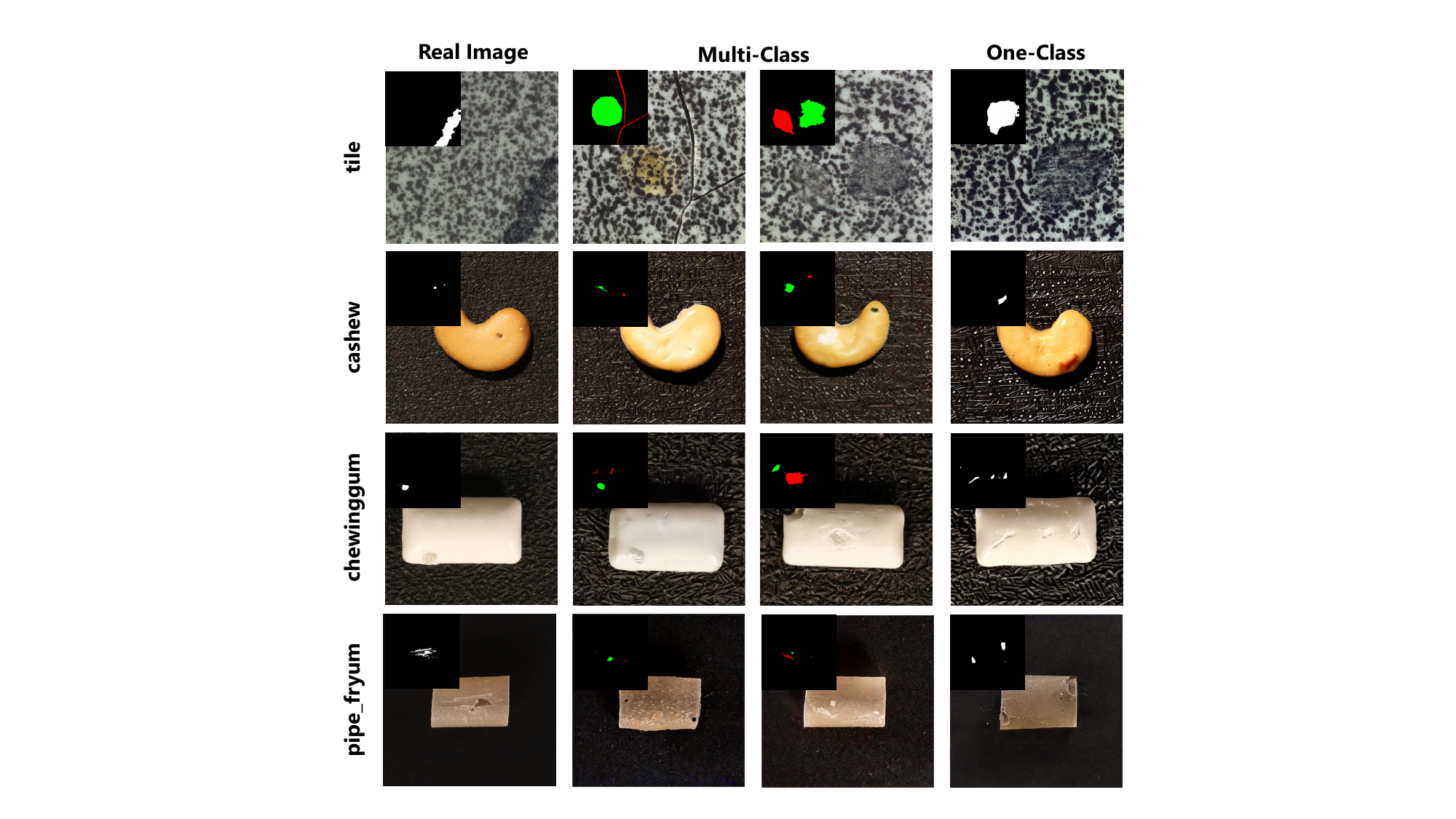}
\caption{Multi-anomaly generation result of GroundingAnomaly.}
\label{fig:mul}
\end{wrapfigure}

\noindent \textbf{Multi-Class anomaly generation.} \noindent A key advantage of GroundingAnomaly is its ability to synthesize multiple, diverse defects of different classes within a single image via grounding anomalies with a semantic map, rather than treating defect combinations as a single rigid \textit{combined} category. The semantic map is constructed by \(S=S_1+S_2\), where \(S_1\) and \(S_2\) represent semantic maps of different anomalies, respectively. And a prompt template 
\[
\begin{gathered}
\texttt{"A photo of a \{PRODUCT\} with} \\
\texttt{\{ANOMALY\(_1\)\} and \{ANOMALY\(_2\)\}"}
\end{gathered}
\]
is employed. As illustrated in \cref{fig:mul}, our model effortlessly generalizes to these complex, multi-class scenarios on MVTec AD and VisA. Additional analyses are available in the supplementary material.

\section{Conclusion} In this paper, we propose GroundingAnomaly, a novel framework for few-shot anomaly synthesis. Unlike prior approaches, our method achieves spatially grounded anomaly generation through a spatial conditioning module, jointly synthesizing anomalies and their host products. Furthermore, a gated self-attention module is introduced to facilitate robust few-shot adaptation. Extensive experiments demonstrate that GroundingAnomaly generates high-quality anomaly images with accurately aligned masks, achieving state-of-the-art performance. In future work, we will explore zero-shot anomaly synthesis and investigate more powerful generative models to achieve improved generation capacity.

\section*{Acknowledgements}
This work was supported in part by the Natural Science Foundation
of Chongqing under Grant CSTB2023NSCQ-MSX0095.

%
%
\bibliographystyle{splncs04}
\bibliography{main}

\clearpage
\setcounter{page}{1}

\appendix
\section{Appendix}
This supplementary material consists of: 
\begin{itemize}
    \item More Implementation Details (\cref{sub:imple}).
    \item More Ablation Studies \& Analysis(\cref{sub:abl}).
    \item More Experiment Results (\cref{sub:res})
\end{itemize}

\subsection{Implementation Details}
\label{sub:imple}
\noindent \textbf{Implementation details of GroundingAnomaly.}
Our method exploits a Stable Diffusion v1.4 \cite{rombach2022high} backbone, the semantic map encoder is a ConvNeXt-Tiny \cite{liu2022convnet} encoder. For disentangled text tokens, we use \(N=4\) and \(K=4\) after ablation. 

All experiments are conducted on NVIDIA RTX 4090 GPU. The AdamW optimizer was employed with a learning rate of \(1 \times 10^{-3}\). The models of all products are trained with a batch size of 4 for 40000 steps. At inference, we utilize 60 steps for sampling and $t'=400$ in NDI. 

\noindent \textbf{Implementation details of anomaly inspection models.}
All generated images are resized to an image size of 256. For U-Net \cite{ronneberger2015u}, we follow the implementation of AnomalyDiffusion \cite{hu2023anomalydiffusion}. For BiSeNet V2 \cite{yu2021bisenet}, a two-branch configuration was adopted: a detail branch comprising three stages with 64, 64 and 128 channels, and a semantic branch of four stages with 16, 32, 64 and 128 channels; the model uses one decode head and four auxiliary heads corresponding to the semantic-branch stages. For UPerNet \cite{xiao2018unified}, ResNet-50 \cite{he2016deep} was adopted as the backbone and the standard UPerNet head (one decode head and one auxiliary head) was used. SegFormer \cite{xie2021segformer} employed the MIT-B0 backbone. For object detection, both Faster R-CNN \cite{ren2015faster} and DETR \cite{carion2020end} use a ResNet-50 backbone (4 stages). For YOLOv5 \cite{glenn_jocher_2020_4154370}, the YOLOv5x was used. 

\noindent \textbf{Metrics.}
Following metrics are employed to evaluate our method:
\begin{itemize}
\item \textbf{Inception Score (IS)} measures image generation quality and diversity by computing the exponential of the Kullback-Leibler (KL) divergence between the marginal label distribution $p(y)$ and conditional label distributions $p(y|x)$ predicted by an Inception-V3 network. Higher IS values indicate superior generation fidelity and diversity.  

\item \textbf{Intra-cluster Pairwise LPIPS Distance (IC-LPIPS)} partitions generated images into $k$ clusters based on LPIPS distance to $k$ target samples, then computes the mean LPIPS distance between images and their assigned cluster's target sample. Higher IC-LPIPS values signify greater generation diversity.  
\item \textbf{Area Under the Receiver Operating Characteristic Curve (AUROC)} quantifies anomaly detection and localization performance by measuring the area under the true positive rate versus false positive rate curve across decision thresholds. Higher AUROC values indicate better performance.  

\item \textbf{Average Precision (AP)} summarizes anomaly detection and localization capability by averaging precision over recall levels on the precision-recall curve. Higher AP values denote superior performance.  

\item \textbf{$F_1$-max} represents the optimal $F_1$ score achieved by threshold selection for binary anomaly classification, balancing precision and recall. Higher $F_1$-max values reflect improved detection and localization accuracy.  

\item \textbf{Maximum Intersection over Union (maxIoU)} measures anomaly localization accuracy by computing the highest IoU between predicted and ground-truth masks across segmentation thresholds. Higher maxIoU values indicate better location capability.  

\item \textbf{Mean Average Precision (mAP)} evaluates object detection performance by averaging average precision (AP) across multiple Intersection-over-Union (IoU) thresholds and object categories. Higher mAP values indicate superior detection accuracy. 
\end{itemize}
IS is computed using the fidelity library and IC-LPIPS using the lpips library. AP, AUROC and F1 are computed with sklearn.metrics. IoU and mAP are evaluated using the default procedures provided by MMSegmentation, MMDetection and the YOLO implementation, respectively. 

\subsection{More Ablation Study \& Analysis}
\label{sub:abl}
\noindent \textbf{Ablation on Disentangled Token Learning.}
We ablate our disentangled tokens with different \(N\) and \(K\) for product tokens \(\{\langle\text{pro}_n\rangle\}_{n=1}^N\) and anomaly tokens \(\{\langle\text{ano}_k\rangle\}_{k=1}^K\). Results are given in \cref{tab:dtl}.

\begin{table}[htbp]
  \centering
  \scriptsize
  \setlength{\tabcolsep}{4pt}
   \caption{Ablation on different numbers of tokens.}
  \resizebox{0.75\columnwidth}{!}{
  \begin{tabular}{c|
    >{\centering\arraybackslash}p{1cm}
    >{\centering\arraybackslash}p{1cm}
    >{\centering\arraybackslash}p{1cm}
    >{\centering\arraybackslash}p{1cm}
    >{\centering\arraybackslash}p{1cm}}
    \toprule
    Model
      & IS & IC-L & AUC-P & AP-P & AP-I \\
    \midrule
    \(N=1\) \(K=1\)
      & 1.79 & 0.37 & 97.2 & 80.5 & 99.1 \\
    \(N=1\) \(K=4\)
      & 1.77 & 0.39 & 97.7 & 79.7 & 98.9 \\
    \(N=4\) \(K=1\)
      & 1.73 & 0.40 & 95.6 & 78.7 & 98.7\\
    \midrule
    \(N=4\) \(K=4\) (\textbf{Ours})
      & \textbf{1.99} & \textbf{0.41} & \textbf{99.2} & \textbf{85.9} &\textbf{99.2} \\
    \bottomrule
  \end{tabular}
  }
  \label{tab:dtl}
\end{table}

\begin{figure}[htbp] 
\centering 
\includegraphics[width=1.0\columnwidth, trim=4.5cm 2.0cm 4.5cm 2.0cm,clip]{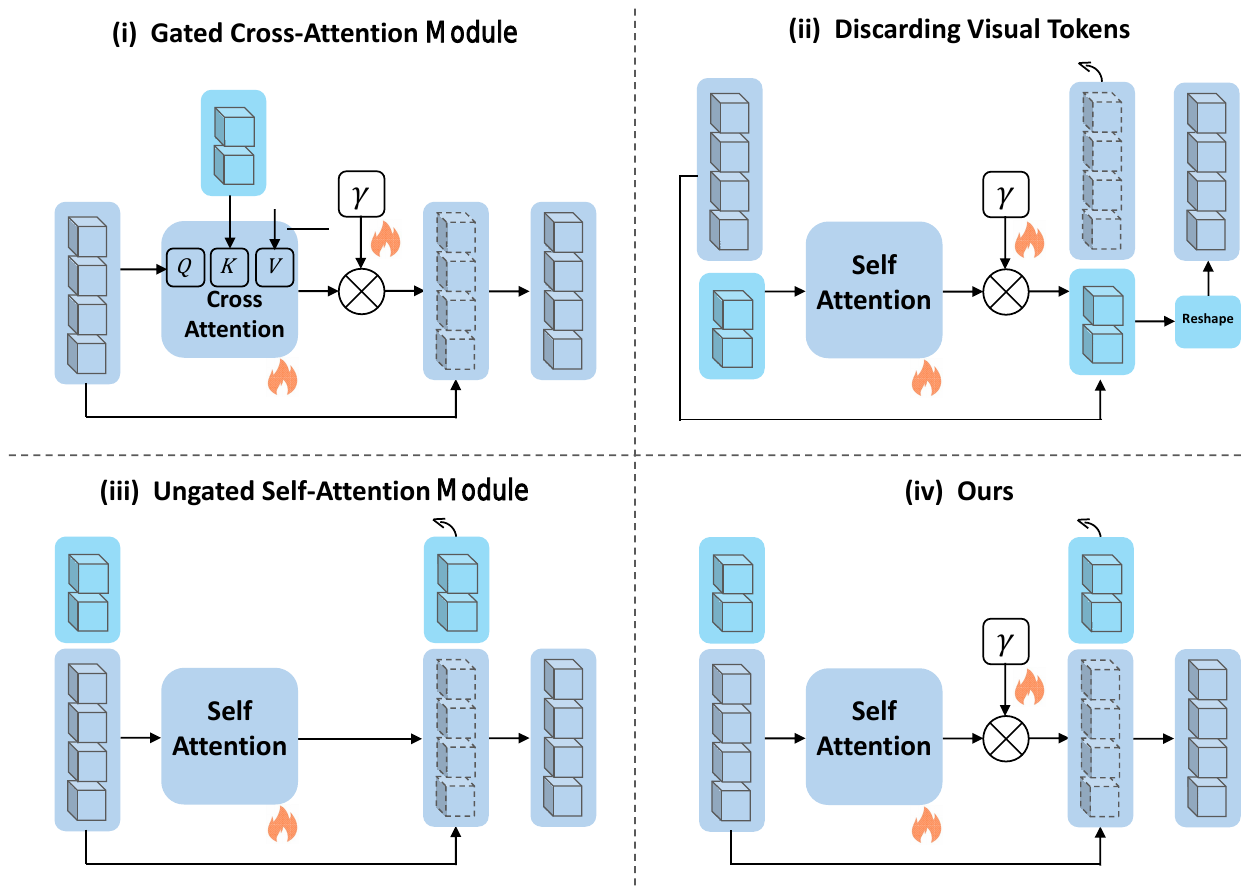} 
\caption{Illustration of ablations on GSM. (i) Gated Cross-Attention Module; (ii) Discarding Visual Tokens; (iii) Ungated Self-Attention Module; (iv) Ours. } 
\label{fig:ill} 
\end{figure}

\noindent \textbf{Ablation on Gated Self-Attention Module.}
We ablate the Gated Self-Attention Module by comparing alternative strategies for injecting conditioning tokens into the U-Net: 
(i) Gated Cross-attention Module (GCM): conditioning tokens are injected via gated cross-attention into the U-Net cross-attention layers; 
(ii) Discarding Visual Tokens (DVT): gated self-attention is applied but visual tokens are replaced by the conditioning tokens, which are reshaped to match the visual-token sequence length.
(iii) Ungated Self-attention Module (USM): conditioning tokens are merged via plain self-attention without a gating scalar;
(iv) Ours without LoRA (w/o LoRA);
(v) Ours: the proposed GSM. The illustration figure is \cref{fig:ill}. 
Experiments are conducted on MVTec AD and results are provided in \cref{tab:gsf}. 

\begin{table}[htbp]
  \centering
  \scriptsize
  \setlength{\tabcolsep}{4pt}
   \caption{Ablation on Gated Self-Attention Module.}
  \resizebox{0.8\columnwidth}{!}{
  \begin{tabular}{c|
    >{\centering\arraybackslash}p{1cm}
    >{\centering\arraybackslash}p{1cm}
    >{\centering\arraybackslash}p{1cm}
    >{\centering\arraybackslash}p{1cm}
    >{\centering\arraybackslash}p{1cm}}
    \toprule
    Model
      & IS & IC-L & AUC-P & AP-P & AP-I \\
    \midrule
    GCM
      & 1.71 & 0.35 & 98.1 & 80.7 & 99.0 \\
    DVT
      & 1.72 & 0.39 & 98.3 & 81.7 & 98.7 \\
    USM
      & 1.68 & 0.40 & 94.9 & 74.1 & 98.0\\
    w/o LoRA
      & 1.93 & 0.39 & 98.3 & 83.3 & 99.0\\
    \midrule
    \textbf{Ours}
      & \textbf{1.99} &\textbf{0.41} & \textbf{99.2} & \textbf{85.9} &\textbf{99.2} \\
    \bottomrule
  \end{tabular}
  }
  \label{tab:gsf}
\end{table}

\begin{figure}[htbp] 
\centering 
\includegraphics[width=0.7\columnwidth, trim=2cm 3cm 2cm 3cm,clip]{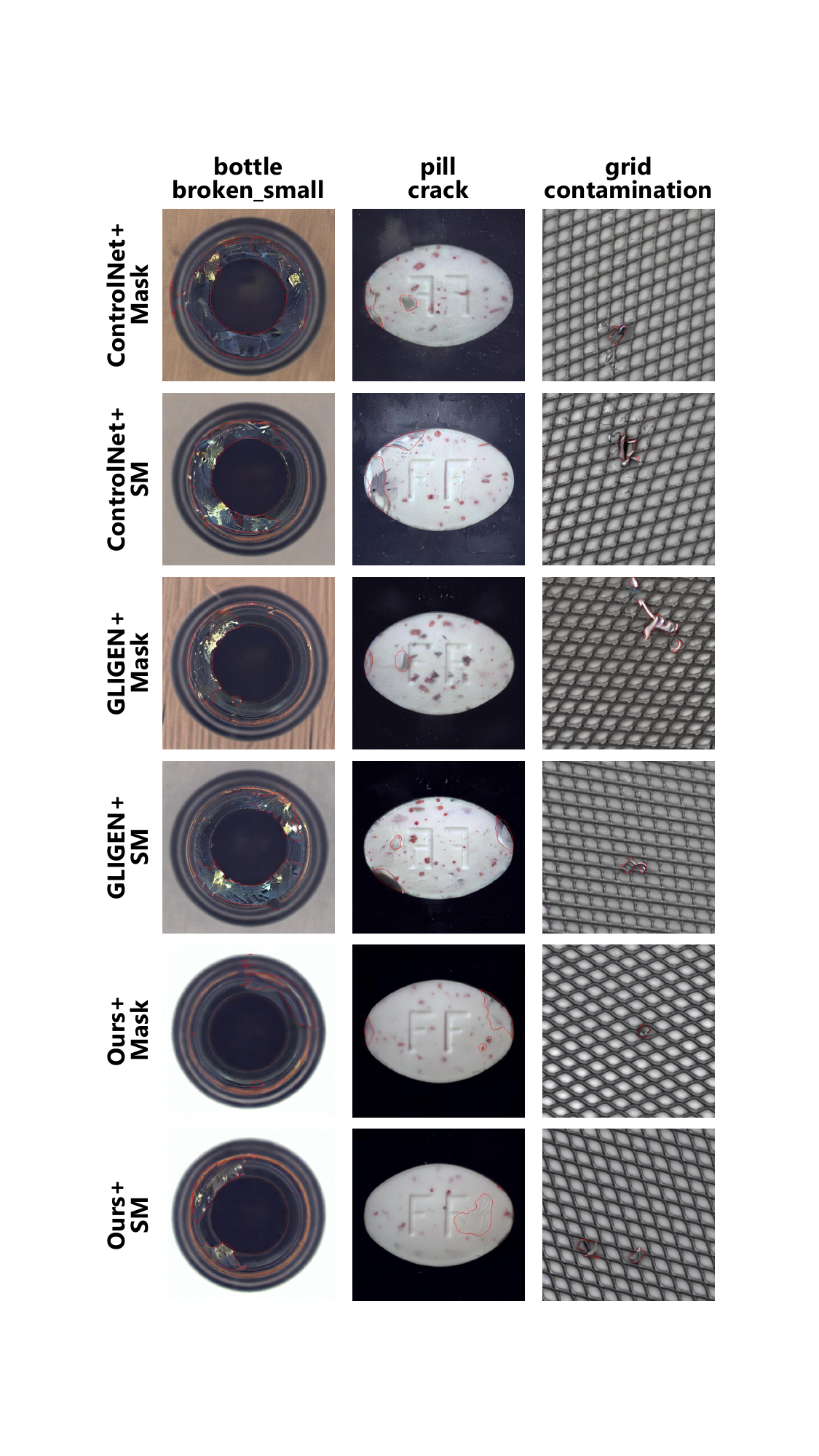} 
\caption{Qualitative comparison of spatial conditioning. GroundingAnomaly achieves the highest fidelity and conditioning accuracy. Furthermore, models conditioned on binary masks, by contrast, often produce anomalies that are less semantically similar to the expected anomaly class.}
\label{fig:spa} 
\end{figure}

\noindent \textbf{Ablation on spatial conditioning.} 
To validate the spatial grounding paradigm of GroundingAnomaly, we conduct a comprehensive ablation study comparing our architecture with controllable diffusion baselines ControlNet \cite{zhang2023adding} and GLIGEN \cite{li2023gligen} using two conditioning signals: class-agnostic binary masks and the proposed multi-class semantic maps (SM). We implement ControlNet as a trainable U-Net encoder copy with zero-convolutions and GLIGEN via self-attention layers copied from the pretrained U-Net. Qualitative results are shown in \cref{fig:spa}, and quantitative results are reported in \cref{tab:spatial}. 

Quantitatively, a consistent ranking is observed: Ours > GLIGEN > ControlNet. ControlNet overfits in the few-shot regime because its large trainable encoder requires substantial data, resulting in degraded fidelity and weak mask adherence. GLIGEN better preserves pretrained priors but still shows inferior fidelity and weaker spatial grounding when adapted to dense conditioning. By contrast, combining the Spatial Conditioning Module (SCM) with the Gated Self-Attention Module (GSM) yields the best mask-to-defect alignment and image quality. Finally, multi-class semantic maps consistently outperform binary masks across methods, since they bind category identifiers to spatial locations to provide rich semantic information and resolve semantic ambiguity.

\begin{table}[htbp]
  \centering
  \scriptsize
  \setlength{\tabcolsep}{4pt}
   \caption{Ablation on spatial conditioning.}
  \resizebox{0.8\columnwidth}{!}{
  \begin{tabular}{c|
    >{\centering\arraybackslash}p{1cm}
    >{\centering\arraybackslash}p{1cm}
    >{\centering\arraybackslash}p{1cm}
    >{\centering\arraybackslash}p{1cm}
    >{\centering\arraybackslash}p{1cm}}
    \toprule
    Model
      & IS & IC-L & AUC-P & AP-P & AP-I \\
    \midrule
    ControlNet+Mask
      & 1.81 & 0.39 & 96.7 & 69.2 & 96.1 \\
    ControlNet+SM
      & 1.86 & 0.40 & 96.9 & 70.3 & 96.2\\
    GLIGEN+Mask
      & 1.85 & 0.39 & 98.1 & 73.3 & 97.7 \\
    GLIGEN+SM
      & 1.88 & \textbf{0.42} & 98.3 & 75.2 & 97.8\\
    Ours+Mask
      & 1.89 & 0.39 & 98.2 & 74.9 & 97.1\\
    \midrule
    \textbf{Ours+SM}
      & \textbf{1.99} & 0.41 & \textbf{99.2} & \textbf{85.9} &\textbf{99.2} \\
    \bottomrule
  \end{tabular}
  }
  \label{tab:spatial}
\end{table}


\noindent \textbf{Generalization to unified segmentation.} 
To further validate the utility of our generated data in more challenging, real-world scenarios, we train unified segmentation models (BiSeNet V2 \cite{yu2021bisenet}, UPerNet \cite{xiao2018unified}, and SegFormer \cite{xie2021segformer}) across all products simultaneously within each dataset. As detailed in \cref{tab:segmentation,tab:image}, incorporating our synthetic image-mask pairs yields consistent performance uplifts, achieving average AP improvements of 3.27\% and 0.60\%, alongside \(F_1\)-max gains of 2.63\% and 0.83\% on MVTec AD and VisA, respectively. 

\begin{table*}[htbp]
  \centering
  \scriptsize
  \setlength{\tabcolsep}{6pt}
  \caption{Comparison on trained unified supervised segmentation models for pixel-level anomaly segmentation on MVTec AD and VisA.}
  \resizebox{\textwidth}{!}{
  \begin{tabular}{l|
                  c c c c|
                  c c c c|
                  c c c c|
                  c c c c|
                  c c c c}
    \toprule
    Model
      & \multicolumn{4}{c|}{DFMGAN}
      & \multicolumn{4}{c|}{AnomalyDiffusion}
      & \multicolumn{4}{c|}{DualAnoDiff}
      & \multicolumn{4}{c|}{SeaS}
      & \multicolumn{4}{c}{Ours} \\
      & AUROC & AP & F1-max & maxIoU
      & AUROC & AP & F1-max & maxIoU
      & AUROC & AP & F1-max & maxIoU
      & AUROC & AP & F1-max & maxIoU
      & AUROC & AP & F1-max & maxIoU \\
    \midrule
    \multicolumn{16}{@{}l}{\textbf{MVTec AD}} \\
    \midrule
    BiSeNet V2
      & 94.57 & 60.42 & 60.54 & 45.83
      & 96.27 & 64.50 & 62.27 & 42.89
      & 96.88 & 66.21 & 61.17 & 46.97
      & 97.21 & 69.21 & 66.37 & \textbf{55.28}
      & \textbf{97.58} & \textbf{75.03} & \textbf{68.71} & 52.34\\
    UPerNet
      & 92.33 & 57.01 & 56.91 & 46.64
      & 96.87 & 69.92 & 66.95 & 50.80
      & 97.17 & 72.08 & 67.55 & 51.15
      & \textbf{97.87} & 74.42 & 70.70 & \textbf{61.24}
      & 97.24 & \textbf{77.09} & \textbf{75.60} & 59.49 \\
    Segformer
      & 93.20 & 58.50 & 58.10 & 47.30
      & 97.10 & 71.20 & 68.00 & 52.10
      & 97.63 & 73.55 & 68.71 & 56.33
      & 97.50 & 81.30 & 74.85 & 59.81
      & \textbf{98.53} & \textbf{82.63} & \textbf{75.51} & \textbf{60.66}\\
    \midrule
    Average
      & 93.37 & 58.64 & 58.52 & 46.59
      & 96.75 & 68.54 & 65.74 & 48.60
      & 97.23 & 70.61 & 65.81 & 51.48
      & 97.53 & 74.98 & 70.64 & \textbf{58.78}
      & \textbf{97.78} & \textbf{78.25} & \textbf{73.27} & 57.50 \\
    \midrule
    \multicolumn{16}{@{}l}{\textbf{VisA}} \\
    \midrule
    BiSeNet V2
      & 75.91 & 9.17 & 15.00 & 9.66
      & 89.29 & 34.16 & 37.93 & 15.93
      & 90.87 & 42.98 & 39.31 & 17.84
      & 96.03 & 42.80 & 45.41 & \textbf{25.93}
      & \textbf{96.11} & \textbf{43.11} & \textbf{45.59} & 24.17\\
    UPerNet
      & 75.09 & 12.42 & 18.52 & 15.47
      & 95.00 & 39.92 & 45.37 & 20.53
      & 95.89 & 53.37 & 49.49 & 24.77
      & 97.01 & 55.46 & 55.99 & 35.91
      & \textbf{97.37} & \textbf{56.69} & \textbf{57.83} & \textbf{36.07}\\
    Segformer
      & 78.33 & 13.47 & 19.59 & 15.13
      & 95.49 & 40.73 & 47.54 & 21.38
      & 95.75 & 50.55 & 48.57 & 22.87
      & 97.58 & 56.39 & 59.41 & 37.88
      & \textbf{97.71} & \textbf{56.65} & \textbf{59.87} & \textbf{37.95}\\
    \midrule
    Average
      & 76.44 & 11.69 & 17.70 & 13.42
      & 93.26 & 38.27 & 43.61 & 19.28
      & 94.17 & 48.97 & 45.79 & 21.83
      & 96.87 & 51.55 & 53.60 & \textbf{33.24}
      & \textbf{97.06} & \textbf{52.15} & \textbf{54.43} & 32.73\\
    \bottomrule
  \end{tabular}
  }
  \label{tab:segmentation}
\end{table*}

\begin{table*}[htbp]
  \centering
  \scriptsize
  \setlength{\tabcolsep}{6pt}
  \renewcommand{\arraystretch}{0.95}
    \caption{Comparison on trained unified supervised segmentation models for image-level anomaly detection on MVTec AD and VisA.}
  \resizebox{\textwidth}{!}{
  \begin{tabular}{l|
                  c c c|
                  c c c|
                  c c c|
                  c c c|
                  c c c}
    \toprule
    Model
      & \multicolumn{3}{c|}{DFMGAN}
      & \multicolumn{3}{c|}{AnomalyDiffusion}
      & \multicolumn{3}{c|}{DualAnoDiff}
      & \multicolumn{3}{c|}{SeaS}
      & \multicolumn{3}{c}{Ours} \\
      & AUROC & AP & F1-max
      & AUROC & AP & F1-max
      & AUROC & AP & F1-max 
      & AUROC & AP & F1-max 
      & AUROC & AP & F1-max  \\
    \midrule
    \multicolumn{16}{@{}l}{\textbf{MVTec AD}} \\
    \midrule
    BiSeNet V2
      & 90.90 & 94.43 & 90.33
      & 90.08 & 94.84 & 91.84
      & 91.33 & 96.79 & 94.49
      & 96.00 & \textbf{98.14} & 95.43
      & \textbf{96.41} & 97.87 & \textbf{95.88} \\
    UPerNet
      & 90.74 & 94.43 & 90.37
      & 96.62 & 98.61 & 96.21
      & 97.89 & 99.00 & 96.77
      & 98.29 & \textbf{99.20} & \textbf{97.34}
      & \textbf{98.63} & 98.77 & 97.24 \\
    Segformer
      & 90.97 & 94.69 & 91.11
      & 95.73 & 98.66 & 96.37
      & 96.97 & 98.79 & 97.03
      & 98.33 & \textbf{99.17} & 97.41
      & \textbf{98.74} & 98.93 & \textbf{97.55} \\
    \midrule
    Average
      & 90.87 & 94.52 & 90.60
      & 94.14 & 97.37 & 94.81
      & 95.40 & 98.19 & 96.10
      & 97.54 & \textbf{98.84} & 96.73
      & \textbf{97.93} & 98.52 & \textbf{96.89} \\
    \midrule
    \multicolumn{16}{@{}l}{\textbf{VisA}} \\
    \midrule
    BiSeNet V2
      & 63.07 & 62.63 & 66.48
      & 76.11 & 77.74 & 73.13
      & 84.77 & 86.72 & \textbf{81.13}
      & 85.61 & 86.64 & 80.49
      & \textbf{88.13} & \textbf{87.14} & 81.03 \\
    UPerNet
      & 71.69 & 71.64 & 70.70
      & 83.18 & 84.08 & 78.88
      & 88.71 & 89.64 & 81.43
      & 90.34 & 90.73 & \textbf{84.33}
      & \textbf{90.81} & \textbf{91.07} & 84.05 \\
    Segformer
      & 72.12 & 71.93 & 71.46
      & 84.77 & 84.53 & 79.54
      & 90.99 & 90.07 & 82.11
      & 91.31 & 90.94 & 85.17
      & \textbf{91.74} & \textbf{91.07} & \textbf{85.73} \\
    \midrule
    Average
      & 68.96 & 68.73 & 69.55
      & 81.35 & 82.12 & 77.18
      & 88.16 & 88.81 & 81.56
      & 89.09 & 89.44 & 83.33
      & \textbf{90.23} & \textbf{89.76} & \textbf{83.60} \\
    \bottomrule
  \end{tabular}
  }
  \label{tab:image}
\end{table*}

\begin{figure}[htbp] 
\centering 
\includegraphics[width=1.0\columnwidth, trim=5cm 4.0cm 5cm 4.0cm,clip]{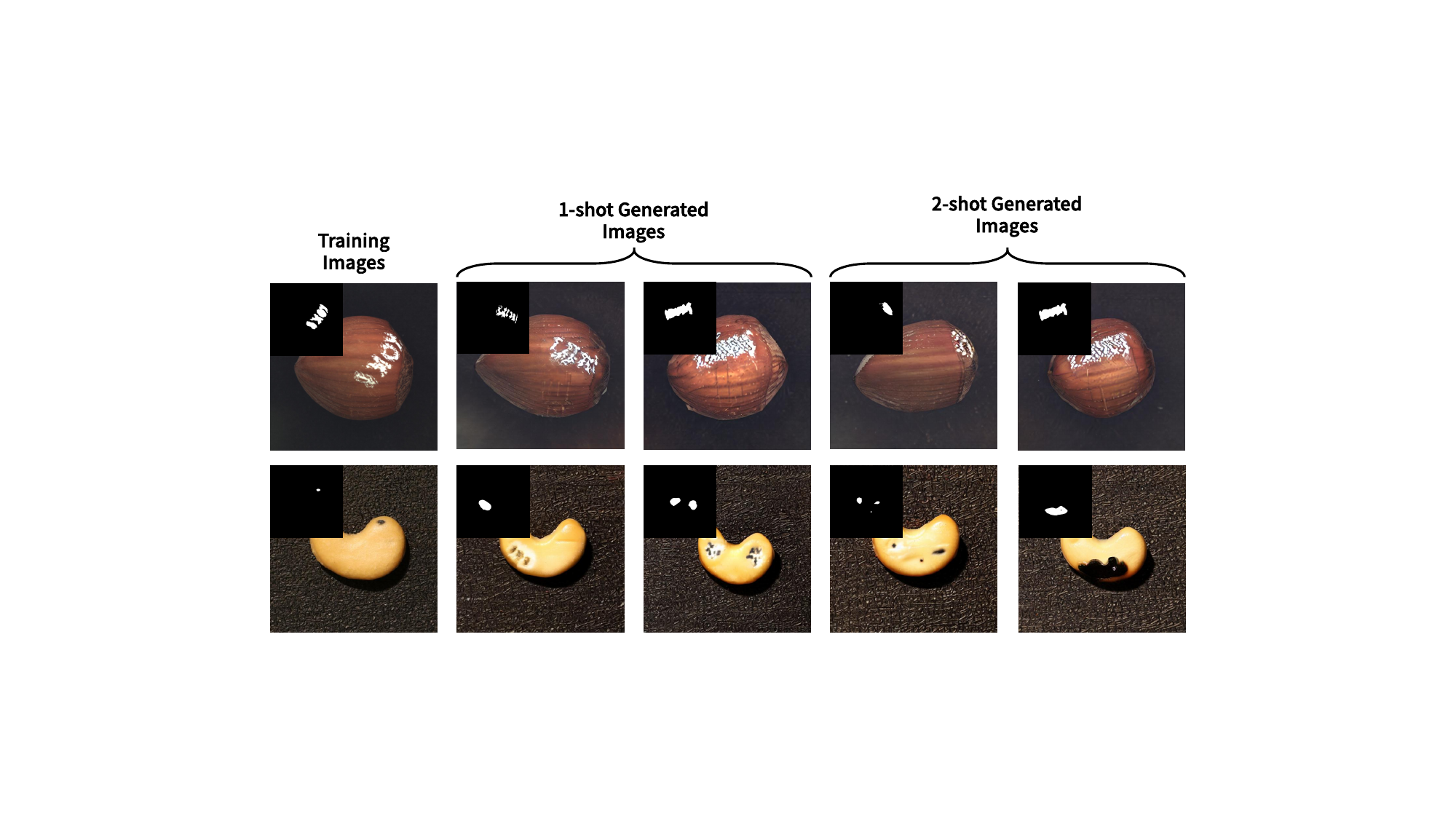} 
\caption{Examples of 1-shot and 2-shot generated images.} 
\label{fig:fewshot} 
\end{figure}

\noindent \textbf{Qualitative results on few-shot generation.} Qualitative results of 1-shot and 2-shot generation are presented in \cref{fig:fewshot}, showing that GroundingAnomaly can generate high-quality anomalous images using only a very small number of real anomaly samples.

\clearpage

\begin{wrapfigure}{r}{0.45\columnwidth}
\centering 
\includegraphics[width=0.45\textwidth, trim=0.0cm 0.0cm 0.0cm 0.0cm,clip]{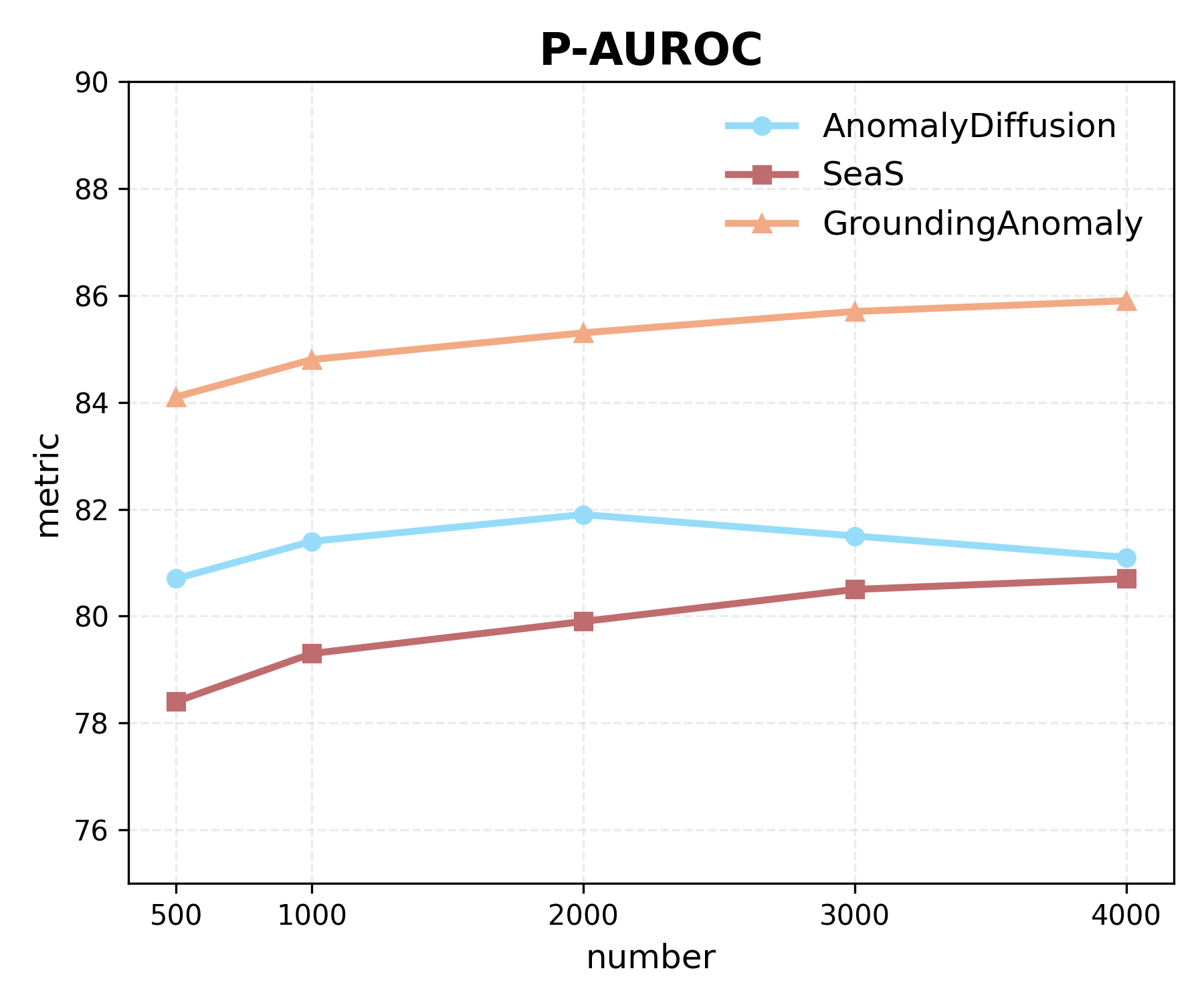} 
\caption{P-AUROC of U-Nets trained by synthesized data.} 
\label{fig:scal} 
\end{wrapfigure}
\noindent \textbf{Analysis on data scaling.} Previous experiments used 1,000 image–mask pairs per anomaly type. To evaluate the effect of scaling the synthesized training set, we generate {500, 1000, 2000, 3000, 4000} images per anomaly type for each synthesis method and train U-Nets on MVTec AD. We report the pixel-level AUROC (P-AUROC) in \cref{fig:scal}, which shows that increasing the amount of GroundingAnomaly-synthesized data consistently enhances downstream inspection performance.

\begin{figure}[htb]
  \centering
  \includegraphics[width=1.0\columnwidth,trim=5cm 6.2cm 5cm 6.2cm,clip]{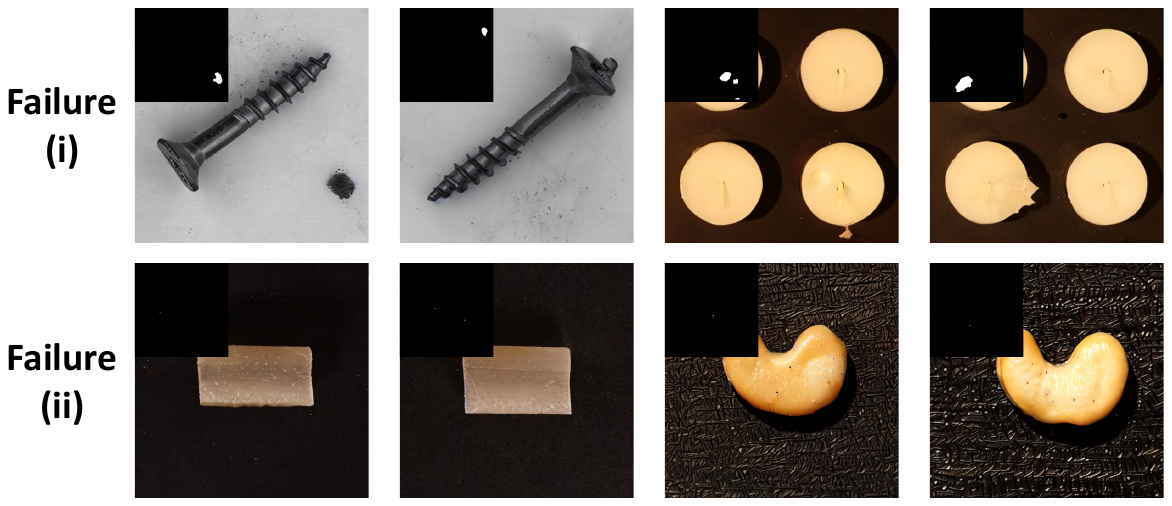}
  \caption{Failure case analysis. }
  \label{fig:failure}
\end{figure}

\noindent \textbf{Analysis on failure cases.} 
Two failure modes are shown in \cref{fig:failure}: (i)~background-region masks causing artifacts outside product boundaries, and (ii)~extremely small anomalies under-resolved in the VAE latent space. These motivate future work on foreground-aware mask generation and stronger backbones with higher resolution.

\subsection{More Experiment Results}
\label{sub:res}
\begin{figure}[htbp] 
\centering 
\includegraphics[width=0.8\columnwidth, trim=3cm 5cm 3cm 5cm,clip]{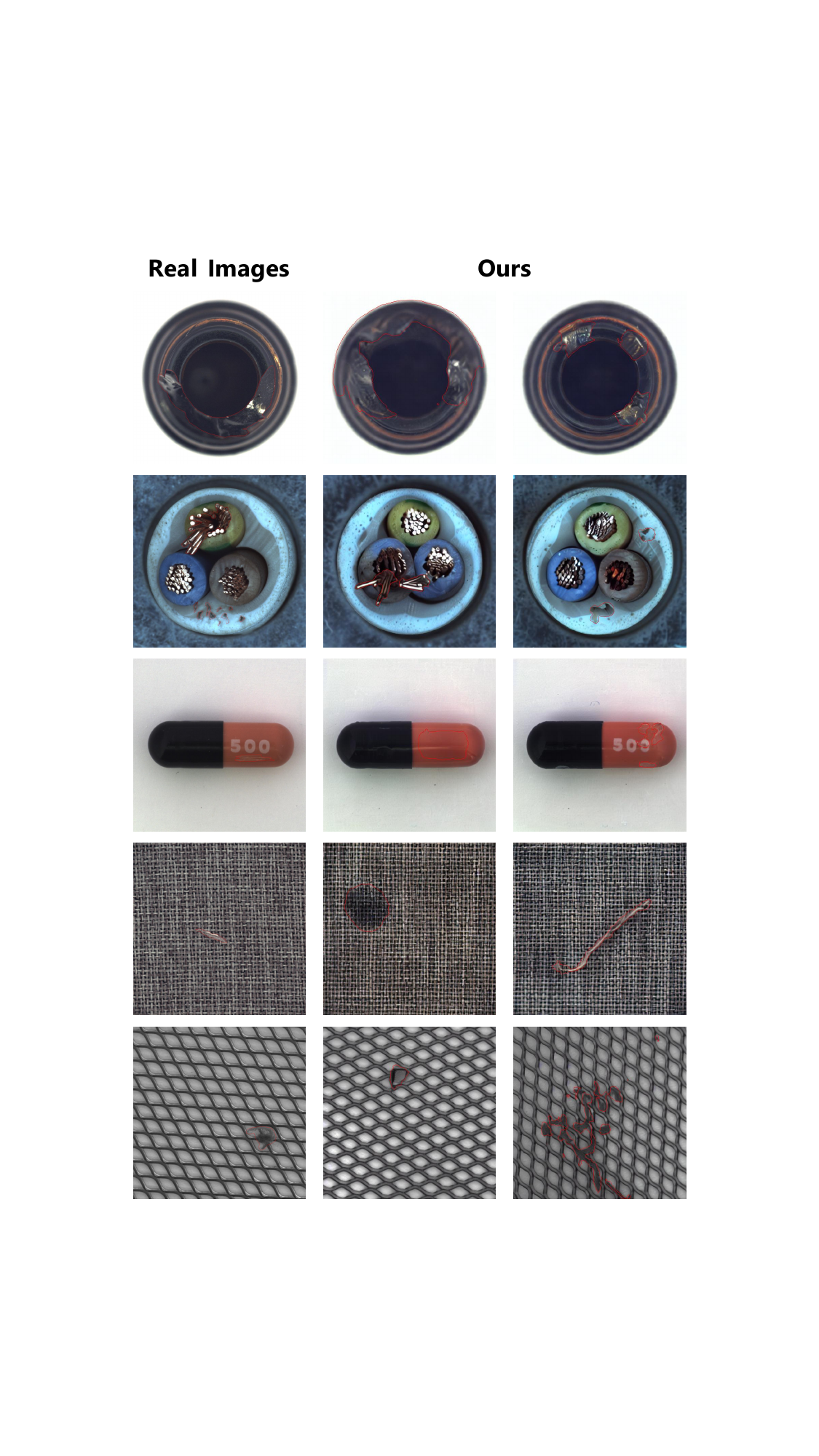} 
\caption{Examples of anomaly generation on MVTec AD.} 
\label{fig:mvtec1} 
\end{figure}

\begin{figure}[htbp] 
\centering 
\includegraphics[width=0.8\columnwidth, trim=3cm 5cm 3cm 5cm,clip]{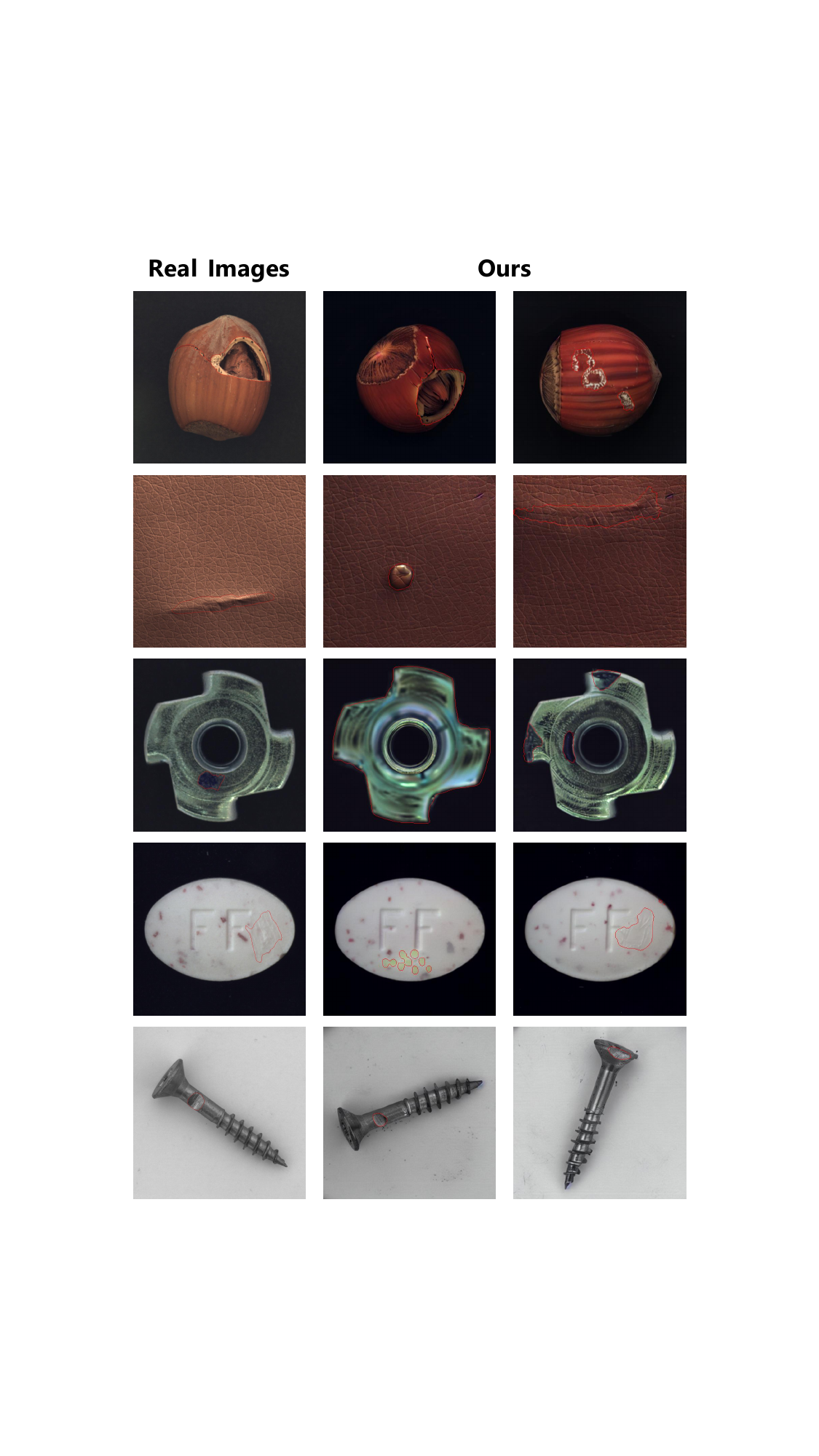} 
\caption{Examples of anomaly generation on MVTec AD.} 
\label{fig:mvtec2} 
\end{figure}

\begin{figure}[htbp] 
\centering 
\includegraphics[width=0.8\columnwidth, trim=3cm 5cm 3cm 5cm,clip]{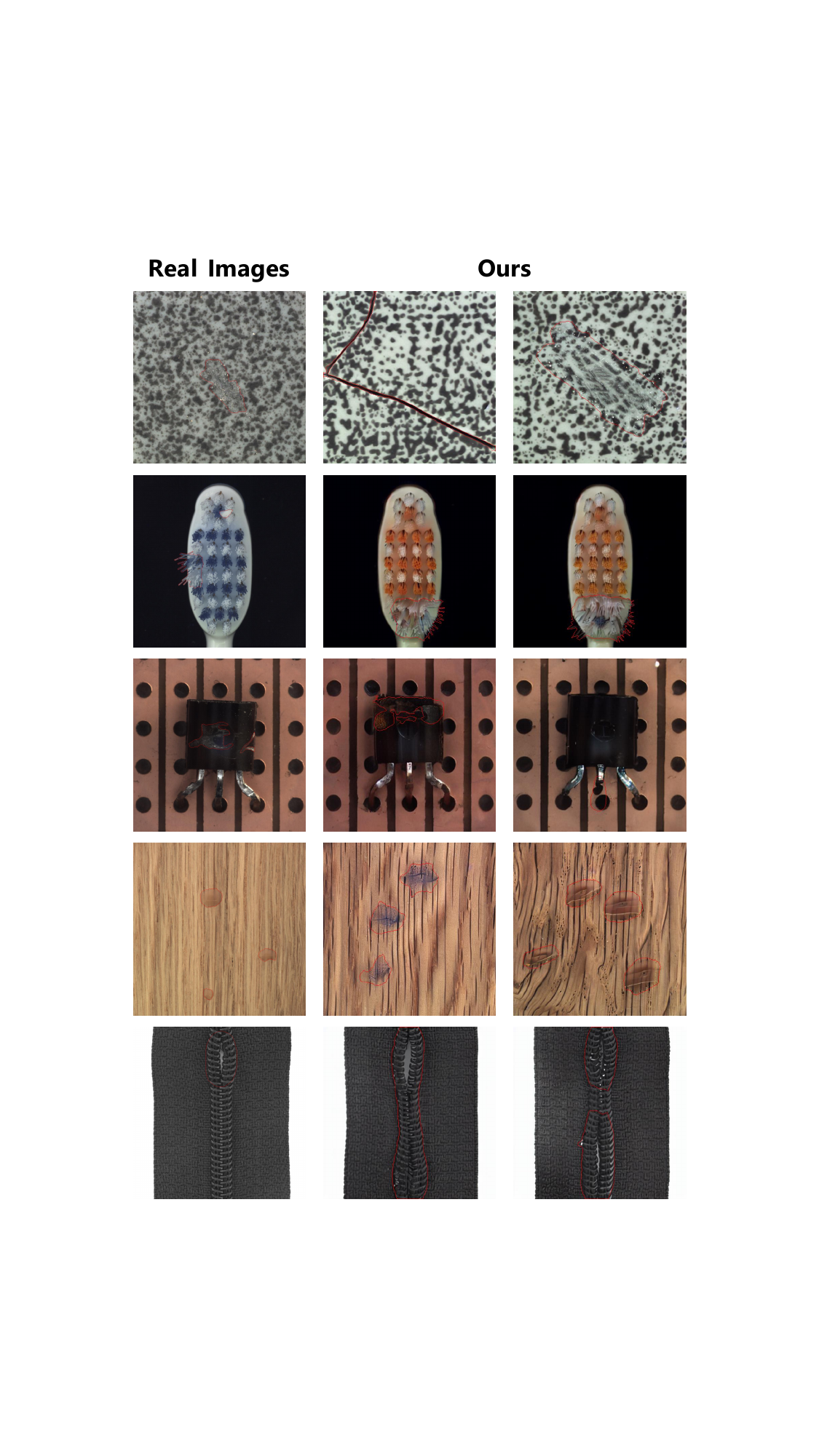} 
\caption{Examples of anomaly generation on MVTec AD.} 
\label{fig:mvtec3} 
\end{figure}

\begin{figure}[htbp] 
\centering 
\includegraphics[width=0.7\columnwidth, trim=3cm 3.5cm 3cm 3.5cm,clip]{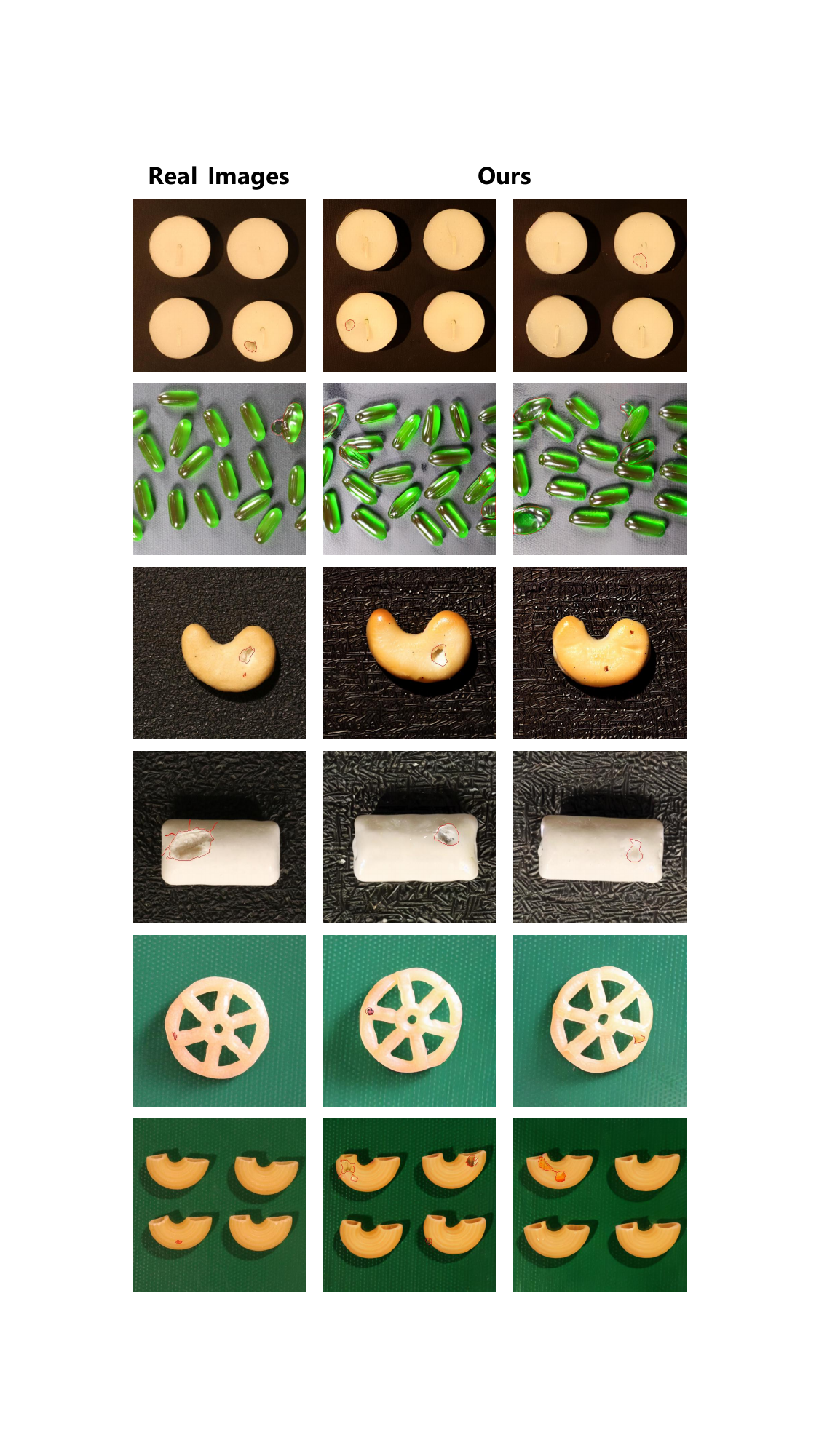} 
\caption{Examples of anomaly generation on VisA.} 
\label{fig:visa1} 
\end{figure}

\begin{figure}[htbp] 
\centering 
\includegraphics[width=0.7\columnwidth, trim=3cm 3.5cm 3cm 3.5cm,clip]{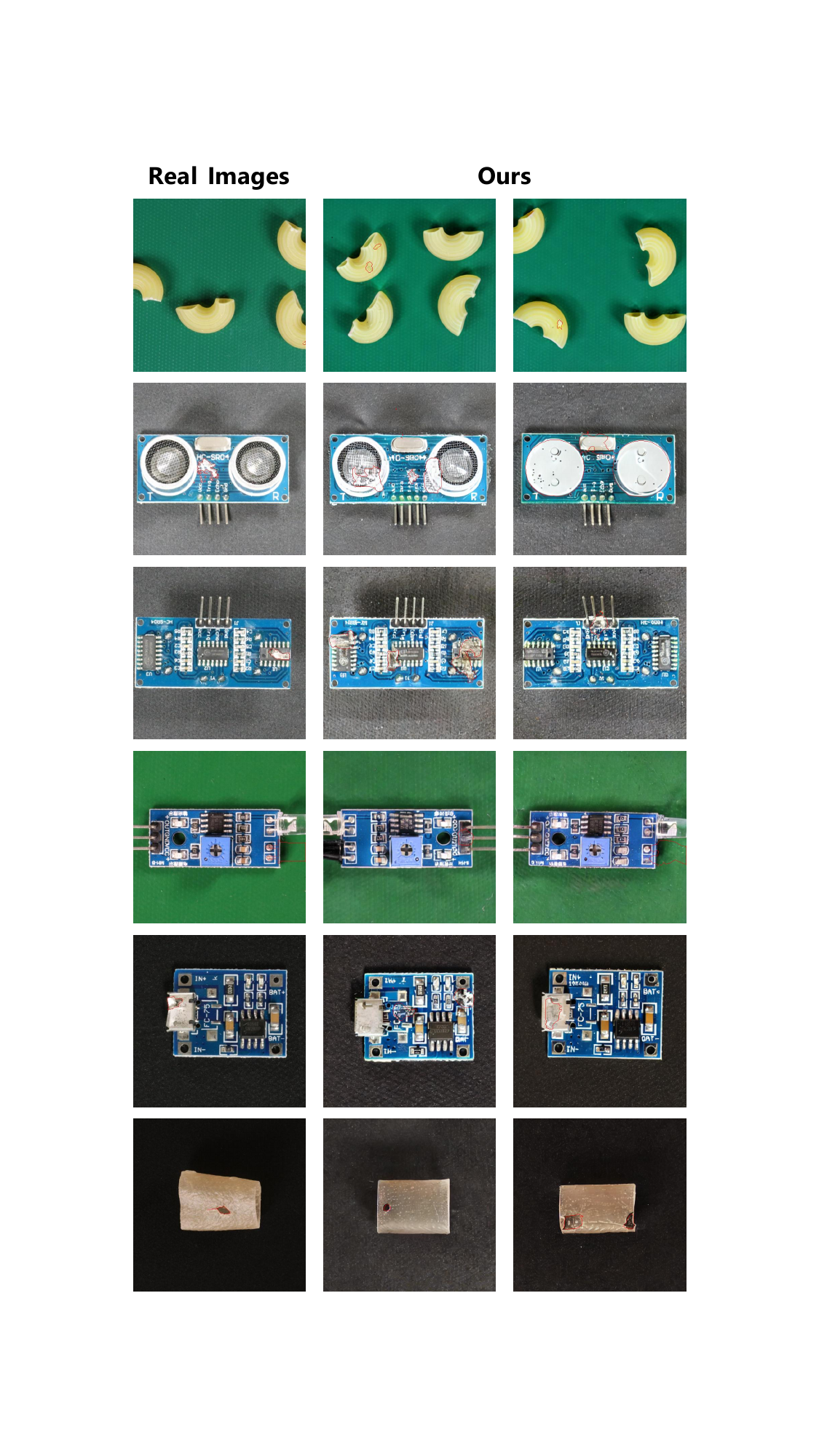} 
\caption{Examples of anomaly generation on VisA.} 
\label{fig:visa2} 
\end{figure}

In this section we present detailed experimental results. More quantitative generation results are provided in \cref{fig:mvtec1,fig:mvtec2,fig:mvtec3,fig:visa1,fig:visa2}, in which the red outlines denote the ground-truth mask. Per-product IS and IC-LPIPS scores are reported in \cref{tab:IS_mvtec,tab:IS_visa}. 

\begin{table*}[t]
  \centering
  \small
  \caption{Per category IS and IC-LPIPS on MVTec AD.}
  \setlength{\tabcolsep}{1.5pt}
  \resizebox{\columnwidth}{!}{
  \begin{tabular}{c|
                     p{1.2cm}<{\centering}p{1.2cm}<{\centering}|
                     p{1.2cm}<{\centering}p{1.2cm}<{\centering}|
                     p{1.2cm}<{\centering}p{1.2cm}<{\centering}|
                     p{1.2cm}<{\centering}p{1.2cm}<{\centering}|
                     p{1.2cm}<{\centering}p{1.2cm}<{\centering}}
    \toprule
    Category 
      & \multicolumn{2}{c|}{DFMGAN} 
      & \multicolumn{2}{c|}{AnoDiff} 
      & \multicolumn{2}{c|}{DualAnoDiff} 
      & \multicolumn{2}{c|}{SeaS} 
      & \multicolumn{2}{c}{\textbf{Ours}} \\
    & IS $\uparrow$ & IC-L $\uparrow$ 
    & IS $\uparrow$ & IC-L $\uparrow$ 
    & IS $\uparrow$ & IC-L $\uparrow$
    & IS $\uparrow$ & IC-L $\uparrow$
    & IS $\uparrow$ & IC-L $\uparrow$ \\
    \midrule
    bottle      & 1.62 & 0.12 & 1.58 & 0.19 & \textbf{2.17} & \textbf{0.36} & 1.78 & 0.21 & 1.97 & 0.35 \\ 
    cable       & 1.96 & 0.25 & 2.13 & 0.41 & 2.12 & \textbf{0.43} & 2.09 & 0.42 & \textbf{2.23} & \textbf{0.43} \\ 
    capsule     & 1.59 & 0.11 & 1.59 & 0.21 & 1.60 & 0.31 & 1.69 & 0.21 & \textbf{1.92} & \textbf{0.34} \\ 
    carpet      & 1.23 & 0.13 & 1.16 & 0.24 & 1.36 & 0.29 & 1.21 & 0.25 & \textbf{1.39} & \textbf{0.30} \\ 
    grid        & 1.97 & 0.13 & 2.04 & 0.44 & 2.09 & 0.42 & 2.62 & 0.44 & \textbf{3.13} & \textbf{0.53} \\ 
    hazelnut   & 1.93 & 0.24 & \textbf{2.13} & 0.31 & 1.91 & 0.35 & 1.89 & 0.31 & 1.85 & \textbf{0.43} \\ 
    leather     & 2.06 & 0.17 & 1.94 & \textbf{0.41} & 1.88 & 0.34 & 2.24 & 0.40 & \textbf{2.51} & 0.35 \\ 
    metal nut   & 1.49 & 0.32 & \textbf{1.96} & 0.30 & 1.56 & 0.32 & 1.68 & 0.31 & 1.75 & \textbf{0.38} \\ 
    pill        & 1.63 & 0.16 & 1.61 & 0.26 & \textbf{1.82} & 0.37 & 1.72 & 0.33 & 1.81 & \textbf{0.55} \\ 
    screw       & 1.12 & 0.14 & 1.28 & 0.30 & 1.34 & \textbf{0.36} & 1.58 & 0.31 & \textbf{1.61} & \textbf{0.36} \\ 
    tile        & 2.39 & 0.22 & 2.54 & \textbf{0.55} & 2.35 & 0.50 & \textbf{2.62} & 0.50 & 1.95 & 0.54 \\ 
    toothbrush  & 1.82 & 0.18 & 1.68 & 0.21 & \textbf{2.40} & \textbf{0.48} & 2.11 & 0.25 & 1.36 & 0.37 \\ 
    transistor  & 1.64 & 0.25 & 1.57 & \textbf{0.34} & \textbf{1.69} & 0.33 & 1.52 & \textbf{0.34} & 1.63 & 0.28 \\ 
    wood        & 2.12 & 0.35 & 2.33 & 0.37 & 2.21 & 0.40 & \textbf{2.78} & \textbf{0.46} & 2.66 & 0.39 \\ 
    zipper      & 1.29 & 0.27 & 1.39 & 0.25 & \textbf{2.09} & 0.36 & 1.69 & 0.30 & 2.05 & \textbf{0.48} \\ 
    \midrule
    Average     & 1.72 & 0.20 & 1.80 & 0.32 & 1.91 & 0.37 & 1.95 & 0.34 & \textbf{1.99} & \textbf{0.41} \\
    \bottomrule
  \end{tabular}}
  \label{tab:IS_mvtec}
\end{table*}

\begin{table*}[t]
  \centering
  \small
  \caption{Per category IS and IC-LPIPS on VisA.}
  \setlength{\tabcolsep}{1.5pt} 
  \resizebox{\columnwidth}{!}{
  \begin{tabular}{c|
                     p{1.2cm}<{\centering}p{1.2cm}<{\centering}|
                     p{1.2cm}<{\centering}p{1.2cm}<{\centering}|
                     p{1.2cm}<{\centering}p{1.2cm}<{\centering}|
                     p{1.2cm}<{\centering}p{1.2cm}<{\centering}|
                     p{1.2cm}<{\centering}p{1.2cm}<{\centering}}
    \toprule
    Category 
      & \multicolumn{2}{c|}{DFMGAN} 
      & \multicolumn{2}{c|}{AnoDiff} 
      & \multicolumn{2}{c|}{DualAnoDiff} 
      & \multicolumn{2}{c|}{SeaS} 
      & \multicolumn{2}{c}{\textbf{Ours}} \\
    & IS $\uparrow$ & IC-L $\uparrow$ 
    & IS $\uparrow$ & IC-L $\uparrow$ 
    & IS $\uparrow$ & IC-L $\uparrow$
    & IS $\uparrow$ & IC-L $\uparrow$
    & IS $\uparrow$ & IC-L $\uparrow$ \\
    \midrule
    candle       & 1.19 & \textbf{0.23} & \textbf{1.28} & 0.17 & 1.27 & 0.15 & 1.20 & 0.12 & 1.25 & 0.17 \\ 
    capsules     & 1.25 & 0.22 & 1.39 & 0.50 & 1.37 & 0.55 & \textbf{1.58} & \textbf{0.60} & \textbf{1.58} & \textbf{0.60} \\ 
    cashew       & 1.25 & 0.24 & 1.27 & 0.26 & \textbf{1.29} & 0.24 & 1.21 & 0.28 & 1.21 & \textbf{0.33} \\ 
    chewinggum   & \textbf{1.33} & 0.24 & 1.15 & 0.19 & 1.19 & 0.20 & 1.29 & 0.27 & 1.29 & \textbf{0.32} \\ 
    fryum        & \textbf{1.28} & 0.20 & 1.20 & 0.14 & 1.17 & 0.14 & 1.14 & 0.21 & 1.14 & \textbf{0.26} \\ 
    macaroni1    & 1.14 & \textbf{0.24} & 1.15 & 0.14 & \textbf{1.17} & 0.16 & 1.15 & 0.18 & 1.15 & 0.23 \\ 
    macaroni2    & 1.47 & 0.38 & 1.56 & 0.38 & 1.60 & 0.39 & 1.57 & 0.39 & \textbf{1.62} & \textbf{0.44} \\ 
    pcb1         & 1.12 & 0.16 & \textbf{1.18} & \textbf{0.35} & 1.17 & 0.32 & \textbf{1.18} & 0.26 & \textbf{1.18} & 0.31 \\ 
    pcb2         & 1.12 & 0.26 & 1.26 & 0.21 & 1.21 & 0.19 & 1.25 & 0.27 & \textbf{1.28} & \textbf{0.32} \\ 
    pcb3         & 1.19 & 0.18 & 1.21 & 0.24 & 1.22 & \textbf{0.26} & 1.22 & 0.21 & \textbf{1.27} & \textbf{0.26} \\ 
    pcb4         & \textbf{1.21} & \textbf{0.28} & 1.14 & 0.25 & 1.13 & 0.21 & 1.15 & 0.22 & 1.15 & 0.27 \\ 
    pipe fryum   & \textbf{1.43} & \textbf{0.32} & 1.29 & 0.17 & 1.27 & 0.18 & 1.31 & 0.16 & 1.36 & 0.21 \\ 
    \midrule
    Average      & 1.25 & 0.26 & 1.26 & 0.25 & 1.25 & 0.25 & 1.27 & 0.26 & \textbf{1.29} & \textbf{0.31} \\
    \bottomrule
  \end{tabular}}
  \label{tab:IS_visa}
\end{table*}

\end{document}